\documentclass[10pt,twocolumn,letterpaper]{article}

\usepackage{cvpr}              % To produce the CAMERA-READY version
\usepackage{graphicx}
\usepackage{amsmath}
\usepackage{amssymb}
\usepackage{booktabs}
\usepackage{nicefrac}
\usepackage{xcolor,colortbl}
\usepackage{multirow}
\usepackage{sidecap}

\newcommand{\red}[1]{\textcolor{red}{#1}}
\newcommand{\blue}[1]{\textcolor{blue}{#1}}

\definecolor{vvlightgray}{rgb}{0.9,0.9,0.9}
\definecolor{vlightgray}{rgb}{0.8,0.8,0.8}

\usepackage[pagebackref,breaklinks,colorlinks]{hyperref}

\usepackage[capitalize]{cleveref}
\crefname{section}{Sec.}{Secs.}
\Crefname{section}{Section}{Sections}
\Crefname{table}{Table}{Tables}
\crefname{table}{Tab.}{Tabs.}

\def\cvprPaperID{7101} % *** Enter the CVPR Paper ID here
\def\confName{CVPR}
\def\confYear{2022}

\begin{document}

\title{Non-isotropy Regularization for Proxy-based Deep Metric Learning}

\author{Karsten Roth$^1$, Oriol Vinyals$^2$, Zeynep Akata$^{1, 3}$\\
$^1$University of Tübingen, $^2$DeepMind, $^3$MPI for Intelligent Systems}
\maketitle

%%%%%%%%%%%%%%%%%%%%%%%%%%%%%%%%%%%%%%%%%%%%%%%%%%%%%%%%%%%%%%%%%%%%%%%%
%%%%%%%%%%%%%%%%%%%%%%%%%%%%%%%%%%%%%%%%%%%%%%%%%%%%%%%%%%%%%%%%%%%%%%%%
\begin{abstract}
Deep Metric Learning (DML) aims to learn representation spaces on which semantic relations can simply be expressed through predefined distance metrics. 
Best performing approaches commonly leverage class proxies as sample stand-ins for better convergence and generalization. 
However, these proxy-methods solely optimize for sample-proxy distances. 
Given the inherent non-bijectiveness of used distance functions, this can induce locally isotropic sample distributions, leading to crucial semantic context being missed due to difficulties resolving local structures and intraclass relations between samples. 
To alleviate this problem, we propose non-isotropy regularization ($\mathbb{NIR}$) for proxy-based Deep Metric Learning. By leveraging Normalizing Flows, we enforce unique translatability of samples from their respective class proxies. This allows us to explicitly induce a non-isotropic distribution of samples around a proxy to optimize for. In doing so, we equip proxy-based objectives to better learn local structures.  
%%%%
Extensive experiments highlight consistent generalization benefits of $\mathbb{NIR}$ while achieving competitive and state-of-the-art performance on the standard benchmarks CUB200-2011, Cars196 and Stanford Online Products. In addition, we find the superior convergence properties of proxy-based methods to still be retained or even improved, making $\mathbb{NIR}$ very attractive for practical usage. Code available at \href{https://github.com/ExplainableML/NonIsotropicProxyDML}{github.com/ExplainableML/NonIsotropicProxyDML}.

% \Zeynep{please try to talk about these constraints a bit. some technical details about the proposed method would help here.}
% % Results
% Extensive experiments highlight consistent improvement of non-isotropic over standard proxy-based DML methods, achieving competitive and state-of-the-art performance on the standard benchmarks CUB200-2011, Cars196 and Stanford Online Products. 
% \Zeynep{any other interesting observation from your ablation studies?}
\end{abstract}

%%%%%%%%%%%%%%%%%%%%%%%%%%%%%%%%%%%%%%%%%%%%%%%%%%%%%%%%%%%%%%%%%%%%%%%%
%%%%%%%%%%%%%%%%%%%%%%%%%%%%%%%%%%%%%%%%%%%%%%%%%%%%%%%%%%%%%%%%%%%%%%%%
\section{Introduction}
\label{sec:intro}
% \Zeynep{I have two generic comments: (1) it is better to use less nouns and more verbs. This usually simplifies readability. For instance 'to inform' is better than 'information'. Nouns like this can not be completely eliminated of course, but as much as possible would already help. (2) sentences spanning more than 2 lines are difficult to follow so I would try to shorten the sentences as much as possible, break them into two if necessary.}
%%% General Introduction
Visual similarity plays a crucial role for applications in image \& video retrieval and clustering \cite{semihard,margin,Brattoli_2020_CVPR}, face re-identification \cite{face_verfication_inthewild,sphereface,arcface} or general supervised \cite{khosla2020supervised} and unsupervised \cite{moco,chen2020simple} contrastive representation learning. A majority of approaches used in these fields employ or can be derived from Deep Metric Learning (DML). DML aims to learn highly nonlinear distance metrics parametrized by deep networks. These networks span a representation space in which semantic relations between images are expressed as distances between respective representations.
\begin{figure}[t!]
    \centering
    \includegraphics[width=0.45\textwidth]{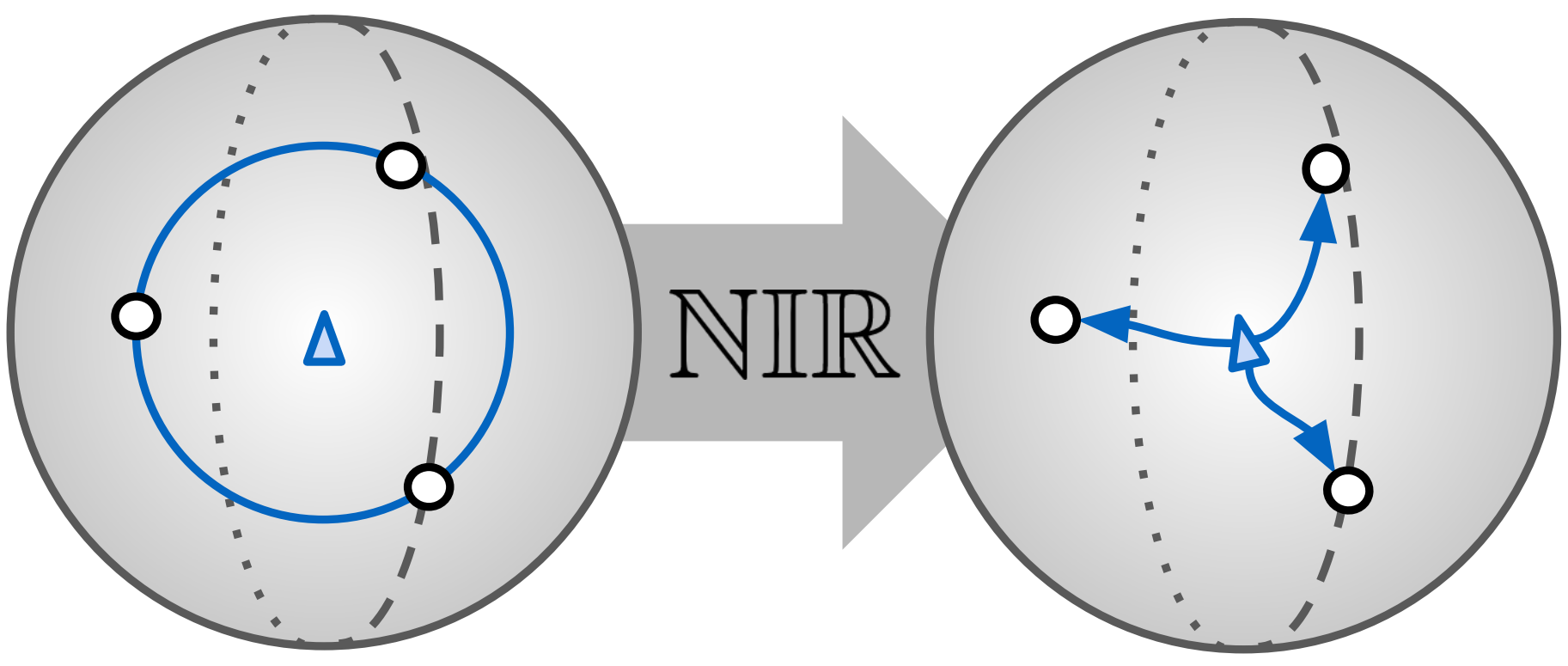}
    \vspace{-3pt}
    \caption{Proxy-based Deep Metric Learning methods optimize for non-bijective similarity measures between proxies ($\blue{\blacktriangle}$) and sample representation ($\circ$), which can introduce local isotropy around proxies, impeding local structures and non-discriminative features to be learned. We propose $\mathbb{NIR}$ to explicitly resolve this.}
    \label{fig:first_page}
    \vspace{-5pt}
\end{figure}
%
%%% Problem Setting
% Methods to learn such representation/embedding spaces are many in number, with various conceptual differences between approaches. 
% A large set of DML methods define ranking tasks over tuples of samples for the network to solve; mapping representation of different samples closer or further apart depending on various constraints such as class labeling. However, these methods rely significantly on 
In the field of DML, methods utilizing proxies have shown to provide among the most consistent and highest performances in addition to fast convergence \cite{proxynca,proxyncapp,kim2020proxy}. 
While other methods introduce ranking tasks over samples for the network to solve, proxy-based methods require the network to contrast samples against a proxy representation, commonly approximating generic class prototypes. Their utilization addresses sampling complexity issues \cite{semihard,margin,smartmining,roth2020pads} inherent to purely sample-based approaches, resulting in improved convergence and benchmark performance. 

However, there is no free lunch. Relying on sample-proxy relations, relations between samples within a class can not be explicitly captured. This is exacerbated by proxy-based objectives optimizing for distances between samples and proxies using non-bijective distance functions. This means, for a particular proxy, that alignment to a sample is non-unique - as long as the angle between sample and proxy is retained, i.e. samples being aligned isotropically around a proxy (see Fig. \ref{fig:first_page}), their distances and respective loss remain the same. This means that samples lie on a hypersphere centered around a proxy with same distance and thus incurring the same training loss. This incorporates an undesired prior over sample-proxy distributions which doesn't allow local structures to be resolved well.
By incorporating multiple classes and proxies (which is automatically done when applying proxy-based losses such as \cite{proxynca,proxyncapp,kim2020proxy,softriple} to training data with multiple classes), this is extended to a mixture of sample distributions around proxies. While this offers an implicit workaround to address isotropy around modes by incorporating relations of samples to proxies from different classes, relying only on other unrelated proxies potentially far away makes fine-grained resolution of local structures difficult. Furthermore, %This issue is made worse 
as training progresses and proxies move further apart. As a consequence, the distribution of samples around proxies, which proxy-based objectives optimize for, comprises modes with high affinity towards local isotropy.
%%%%%
This introduces semantic ambiguity, as semantic relations between samples within a class are not resolved well. 
However, a lot of recent work has shown that understanding and incorporating these non-discriminative relations drives generalization performance \cite{dvml,hardness-aware,mic,roth2020revisiting,milbich2020diva}.

\begin{figure*}[t!]
    \centering
    \includegraphics[width=0.95\textwidth]{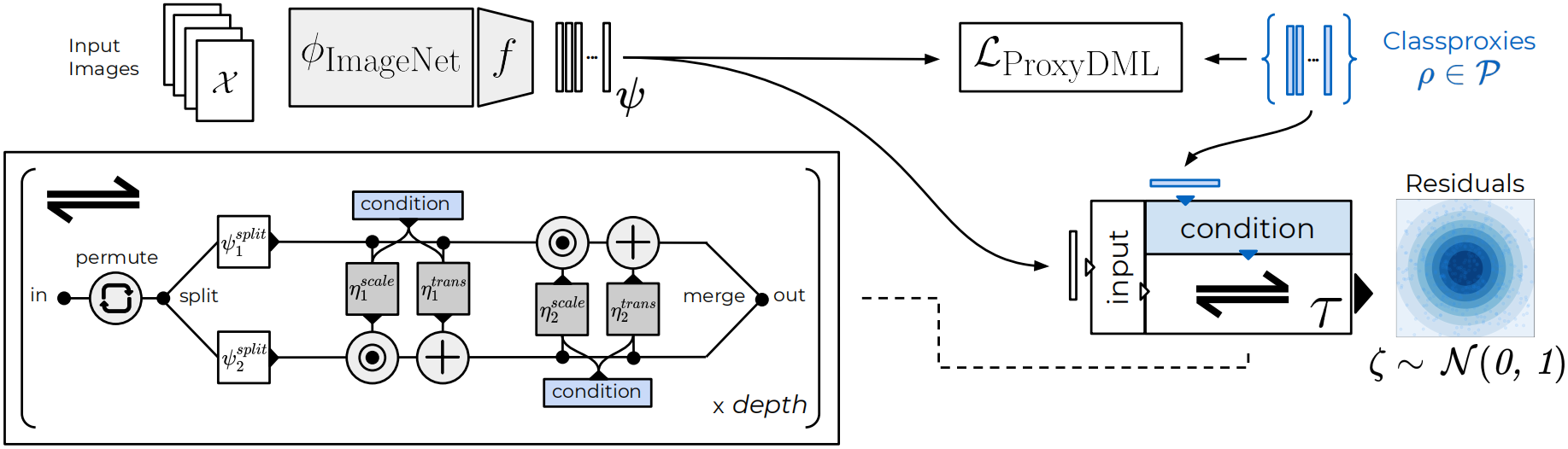}
    \vspace{-5pt}
    \caption{$\mathbb{NIR}$ - \textit{Non-isotropy Regularisation.} We refine the distribution of samples $\psi$ around class proxies $\rho$, $p(\psi|\rho)$, learned via proxy-based DML by leveraging Normalizing Flows ($\tau$, $\leftrightharpoons$). These allow us to define a bijective translation $\tau$ which uses a simple density $q = \mathcal{N}(0, 1)$ of residuals $\zeta$ to induce a distribution over unique sample-proxy relations, $p(\tau(\zeta|\rho)|\rho)$. This allows for better resolution of local structures and non-discriminative features to be learned, improving generalisation while retaining fast convergence.}
    \label{fig:setup}
    \vspace{-5pt}
\end{figure*}

%%% Solution
To tackle this issue without resorting to sample-based objectives that impede the superior convergence and generalization of proxy-based approaches, this work proposes non-isotropy regularisation ($\mathbb{NIR}$) for proxy-based DML. 
$\mathbb{NIR}$ extends proxy-based objectives to encourage explicitly learning unique sample-proxy relations and eliminating semantic ambiguity. %to be removed.
In detail, we introduce a novel uniqueness constraint, in which samples within a class must be uniquely and sufficiently described by a (non-linear) translation from the respective class proxy. This explicitly induces a distribution for proxy-based objectives to match in which isotropy and ambiguity is heavily penalized.
% \oriol{is this true? if you try to make the samples unique / being able to be described by a transformation, wouldn't this precisely push all the samples to fully occupy the ball around the proxy?}
We achieve uniqueness by leveraging a bijective and thus invertible family of translations. As the proxy-sample translations need to adapt to the specific domain at hand, we require both trainability and non-linearity of our translation models. These functional constraints are naturally expressed through Normalizing Flows and Invertible Networks \cite{cinn,nf_1,nf_2}. Using conditional variants, we then formalize $\mathbb{NIR}$ where %by requiring 
sample relations %to be (uniquely) mappable to 
are (uniquely) mapped by a Normalizing Flow given some residual conditioned on the respective class proxy.

Extensive experiments show that $\mathbb{NIR}$ indeed introduces higher feature diversity, reduces overclustering, increases uniformity in learned representation spaces and learns more diverse class-distributions than non-regularized counterparts.
Evaluating our approach on the standard DML benchmarks CUB200-2011 \cite{cub200-2011}, CARS196 \cite{cars196} and Stanford Online Products \cite{lifted} showcases improved generalization capabilities of $\mathbb{NIR}$-equipped proxy DML, achieving competitive or state-of-the-art performance while retaining or even improving convergence speeds.

%%%%%%%%%%%%%%%%%%%%%%%%%%%%%%%%%%%%%%%%%%%%%%%%%%%%%%%%%%%%%%%%%%%%%%%%
%%%%%%%%%%%%%%%%%%%%%%%%%%%%%%%%%%%%%%%%%%%%%%%%%%%%%%%%%%%%%%%%%%%%%%%%
\section{Related Works}
\label{sec:related_works}
Deep Metric Learning (DML) has driven research in image and video retrieval \& zero-shot clustering applications \cite{semihard,npairs,margin,Brattoli_2020_CVPR}, with particular applications for example in person re-identification \cite{face_verfication_inthewild,sphereface,arcface,cosface} and as an auxiliary tool for improved supervised \cite{khosla2020supervised} and unsupervised representation learning \cite{moco,chen2020simple,pretextmisra}. Commonly proposed DML methods introduce ranking tasks for networks to solve as training surrogates. These can involve ranking constituents in tuples (such as pairs \cite{contrastive,musgrave2020metric}, triplets \cite{semihard,face_verfication_inthewild,margin} or higher-order tuples \cite{npairs,lifted,multisimilarity}) to contrasting between sample and prototypical representations \cite{proxynca,proxyncapp,kim2020proxy,softriple,Zhu2020graphdml}. These prototypical- or proxy-based approaches are commonly introduced in order to address sampling complexity issues when sampling tuples for a network to solve, which are otherwise addressed through various sampling heuristics \cite{semihard,margin,smartmining,roth2020pads,htl}. 
We propose a natural extension to proxy-based approaches by addressing a major shortcoming introduced when only contrasting between samples and proxies while retaining beneficial properties of these methods.
Finally, recent work has focused on generic extensions to DML to improve the quality of learned representation spaces through divide-and-conquer \cite{Sanakoyeu_2019_CVPR}, synthetic data \cite{dvml,hardness-aware}, adversarial and graph-based training \cite{daml,seidenschwarz2021graphdml,Zhu2020graphdml}, bypassing representation bottlenecks \cite{horde,s2sd}, attention \cite{abe} and auxiliary or few-shot feature learning \cite{mic,milbich2020sharing,roth2020revisiting,milbich2020diva,milbich2021characterizing,dullerud2022is}. These works offer distinct, orthogonal benefits.

%%%%%%%%%%%%%%%%%%%%%%%%%%%%%%%%%%%%%%%%%%%%%%%%%%%%%%%%%%%%%%%%%%%%%%%%
%%%%%%%%%%%%%%%%%%%%%%%%%%%%%%%%%%%%%%%%%%%%%%%%%%%%%%%%%%%%%%%%%%%%%%%%
\section{Non-isotropic Deep Metric Learning}
\label{sec:methods}
% This section introduces preliminaries (\S\ref{subsec:preliminaries}) before motivating (\S\ref{subsec:uniqueness}) and introducing the proposed non-isotropy regularisation $\mathbb{NIR}$ in \S\ref{subsec:nir}.

%\subsection{Preliminaries}
%\label{subsec:preliminaries}
A DML model defines a distance metric $d_\psi(x_i, x_j)$ over images $x_i\in\mathcal{X}$ parametrized by a feature extraction backbone $\phi$ and a projection $f$ onto the final metric space $\Psi\subset\mathbb{R}^d$, such that $\Psi := f\circ \phi (\mathcal{X})$. $\Psi$ is commonly normalized to the unit hypersphere \cite{margin,wang2020understanding} such that $\Psi = S^{d-1}_\Psi$. This metric space is commonly equipped with a predefined distance metric such as the euclidean distance $d(\bullet, \bullet)$ or cosine similarity $s(\bullet, \bullet)$, which are equivalent on the hypersphere \cite{eq_1,eq_2}. 
%%%%%%%
During training, DML learns $\psi = f\circ\phi$ to connect $d_\psi(x_i, x_j) := d(\psi_i, \psi_j)$ to the semantic similarity of images $x_i$ and $x_j$. Training methods commonly involve the definition of ranking tasks for the network to solve - given e.g. a triplet of anchor $x_a$, positive $x_p$ and negative $x_n$ with $y_a = y_p \neq y_n$ where $y\in\mathcal{Y}$ denotes the respective class
% , the common triplet loss can be formulated as
% \begin{equation}
% \begin{split}
%     \mathcal{L}_\text{triplet} = \frac{1}{|\mathcal{T}_\mathcal{B}|} \sum_{\mathcal{T}_\mathcal{B}} \left[d_\psi(x_a, x_p) - d_\psi(x_a, x_n) + \gamma\right]_+
% \end{split}
% \end{equation}
% with margin $\gamma$ 
and triplets  $(x_a,x_p,x_n)\in\mathcal{T}_\mathcal{B}$ sampled from a minibatch $\mathcal{B}$. 
%%%%
However, tuple sampling is difficult, as the tuple space complexity increases with tuple dimensionality; incurring a lot of redundancy \cite{semihard,margin,smartmining}. While sampling heuristics have been introduced to address this \cite{semihard,margin,smartmining,roth2020pads}, recent work \cite{proxynca,proxyncapp,kim2020proxy} has heuristically supported the promise of proxy-based ranking objectives.

\subsection{On the shortcoming of proxy-based DML}
\label{subsec:uniqueness}
% \Zeynep{although this part makes the paper self contained, there seems to be too much information about a prior technique. I would move most of this subsection to the supplementary and keep only the essentials necessary for the upcoming subsection focusing on your technique. Also the heavy notation makes it difficutl to follow. Section 3.2 should be the main emphasis here, e.g. similar to the model figure, where ProxyDML is a small branch which doesn't contain much details, I would reduce its load here from the text as well. }\oriol{agree we can simplify it a bit by showing only eq 2 without the extra symbols (softmax), and then state in text variations.}
Proxy-based objectives use contrastive operations not between samples (e.g. via the cosine similarity $s_\psi(x_i, x_j)$ \cite{multisimilarity,kim2020proxy,arcface}), but between class-prototypical (class proxies) representations $\rho_j\in\mathcal{P}$ $s(\psi_i, \rho_j)$ for classes $y_i$ and $y_j$. This removes the need for complex sampling operations in methods reliant on sample-based tuples, which allows proxy-based objectives to benefit from fast convergence and good generalization performance. 

However, this property also incurs the strongest shortcoming, because relying on sample-proxy pairs and the non-bijective similarity measure $s(\psi, \rho):= s(\psi_i, \rho_{y_{\psi_i}})$ can induce features to locally follow an isotropic distribution around the proxy. 
This can be seen more explicitly when looking at the sample-proxy distributions various proxy-objectives optimize for.
Take for example the foundational ProxyNCA objective \cite{proxynca}. ProxyNCA is heavily connected to various recent, state-of-the-art objectives (such as ProxyAnchor \cite{Kim_2021_CVPR} or SoftTriple \cite{softriple}) and has the form
\begin{equation}
\label{eq:pnca}
    \mathcal{L}_\text{PNCA} = -\mathbb{E}_{\substack{x\sim\mathcal{X}_y\\y\sim\mathcal{Y}}}\left[ \log \left( \frac{e^{s(\psi(x), \rho_y)}}{\sum_{\rho^*\in\mathcal{P}^{-y}} e^{s(\psi(x), \rho^*)}} \right) \right]
\end{equation}
with the complete set of class proxies $\mathcal{P}$ and with class $y$ removed $\mathcal{P}^{-y}$\footnote{We use cosine similarity instead of the euclidean distance as done in \cite{proxynca,proxyncapp}, as both are equivalent on the hypersphere.} that are trained jointly during training. 
Minimizing the distance of samples to their respective class proxies while maximizing it for non-related proxies, this objective can be regarded as implicitly maximizing the log-likelihood of samples $\psi$ belonging to proxy $\rho$ (such that $y_\psi = y_\rho$) under a von-Mises-Fisher (vMF\footnote{An assumption found also e.g. in self-supervised learning \cite{dml_inversion}.}) mixture model \cite{vmf_mix_1,vmf_mix_2} around directions $\rho$
\begin{align}
\label{eq:vmfmm}
    p_\text{vMFmm}(\psi|\rho) &= \frac{\pi_\rho C_d(\kappa_{\rho})e^{ \kappa_{\rho} s(\psi, \rho)}}{\sum_{\rho^*\in\mathcal{P}} \pi_{\rho^*} C_d(\kappa_{\rho^*}) e^{\kappa_{\rho^*} s(\psi, \rho^*)}}\\
    C_d(\kappa) &= \kappa^{d/2-1} \cdot \left[(2\pi)^{d/2}I_{d/2-1}(\kappa)\right]^{-1}
\end{align}
assuming a class-independent concentration parameter $\kappa_\rho = \kappa$ and mixture $\pi_\rho = \pi$ such that $C_d(\kappa_\rho) = C_d(\kappa) = const$\footnote{$C_d$ incorporates the modified Bessel function $I_p$ of the first kind and order $p$, which can be neglected here as $C_d$ cancels out}. 
Even more, \cite{proxyncapp} show that performance of $\mathcal{L}_\text{PNCA}$ improves when actually optimize the proxy-assignment probability (by replacing $\mathcal{P}^{-y}$ with $\mathcal{P}$ in the denominator, giving $\mathcal{L}_\text{PNCA++}$) directly, which equals an explicit negative log-likelihood minimization of $p_\text{vMFmm}$:
\begin{equation}
\label{eq:pncapp}
    \mathcal{L}_\text{PNCA++} = -\mathbb{E}_{x\sim\mathcal{X}_y, y\sim\mathcal{Y}}\left[\log p_\text{vMFmm}(\psi(x)|\rho_y)\right]
\end{equation}

% In a similar fashion, recent and state-of-the-art proxy objectives that extend upon ProxyNCA, such as the ProxyAnchor objective
% \begin{equation}
%     \mathcal{L}_\text{PA} = \frac{1}{|\mathbb{P}^+|} \sum_{\rho\in\mathbb{P}^+} \log \left( 1 + \sum_{x\in\mathcal{B}, y_x = y_\rho} e^{-\alpha \cdot [s(x, \rho) - \delta]}\right) + \frac{1}{|\mathbb{P}|} \sum_{\rho\in\mathbb{P}} \log \left( 1 + \sum_{x\in\mathcal{B}, y_x \neq y_\rho} e^{\alpha \cdot [s(x, \rho) + \delta]}\right)
% \end{equation}
% while claiming to better incorporate ``data-data''-relations \cite{kim2020proxy}, similarly only rely on sample-proxy relations to learn a respective metric space. 

% \red{Ideally, we can derive the ProxyAnchor objective from $p_\text{vMFmm}$ or something similar.}
Recent and state-of-the-art proxy objectives that extend upon ProxyNCA, such as the ProxyAnchor \cite{kim2020proxy} objective
\begin{equation}
\label{eq:panchor}
\begin{split}
    \mathcal{L}_\text{PA} = &\frac{1}{|\mathcal{P}^+|} \sum_{\rho\in\mathcal{P}^+} \log \left( 1 + \sum_{x\in\mathcal{B}, y_x = y_\rho} e^{-\alpha \cdot [s(x, \rho) - \delta]}\right) \\&+ \frac{1}{|\mathcal{P}|} \sum_{\rho\in\mathcal{P}} \log \left( 1 + \sum_{x\in\mathcal{B}, y_x \neq y_\rho} e^{\alpha \cdot [s(x, \rho) + \delta]}\right)
\end{split}
\end{equation}
operate under similar assumptions, suggesting slight, more hyperparameter-heavy variations to the loss terms. While ProxyAnchor specifically suggests pulling samples towards proxies instead of proxies towards samples as done in $\mathcal{L}_\text{PNCA}$, it similarly relies on the same sample-proxy contrastive operations to learn a respective metric space.
This means that these methods can only learn sample distributions around proxies that are closely related to $p_\text{vMFmm}$-like distributions learned via $\mathcal{L}_\text{PNCA}$ .
% even when claiming to better incorporate ``data-data''-relations \cite{kim2020proxy} through the inverted contrastive operation. 
Indeed, our experiments (see \S\ref{subsec:exp_nir} and Tab. \ref{tab:relative_results}) show that when adapting the ProxyNCA objective to the ProxyAnchor formulation
\begin{equation}
\label{eq:pnca_ada}
\begin{split}
    &\mathcal{L}_\text{PNCA}^* = \frac{1}{|\mathbb{B}^+|} \sum_{x\in\mathcal{B}} \log \left( 1 + e^{-\alpha \cdot [s(x, \rho_{y_x}) - \delta]}\right) \\&+ \frac{1}{|\mathcal{B}|} \sum_{x\in\mathcal{X}} \log \left( 1 + \sum_{\rho\in\mathcal{P}, y_x \neq y_\rho} e^{\alpha \cdot [s(x, \rho) + \delta]}\right)
\end{split}
\end{equation}
performance becomes much more similar than indicated in \cite{kim2020proxy}. And while ProxyAnchor may not optimize for the exact $p(\psi|\rho)$ formulation, the results support a very strong distributional relation between these proxy-objectives.

However, mixture distributions such as $p_\text{vMFmm}$ suffer from several issues. Firstly, each mode on its own is isotropic as $s(\psi, \rho)$ returns the same value as long as the angle $\theta(\psi, \rho)$ is retained. 
This means that class-specific structures can only be resolved implicitly through relations with proxies of different classes. Secondly, this intraclass resolution becomes worse as training progress, since proxies from different classes continue to contrast further and sample-proxy relations for same-class pairs are overshadowed (see e.g. Eq. \ref{eq:pnca}, \ref{eq:panchor}, \ref{eq:pnca_ada}). Similarly, for samples closer towards each respective class proxy, resolving local structures becomes harder. Effectively, this results in learned sample distributions to have a strong affinity towards local isotropy.

%%%
As such, proxy-based objectives are inherently handicapped in resolving local intraclass clusters and structures. Consequently, semantic relations between samples within a class are not well encoded in the representational structure of the learned deep metric spaces. 
However, the ability to explain and represent intraclass sample relations has been consistently shown to be a key driver for downstream generalization performance in DML \cite{milbich2020diva,mic,dvml,milbich2020sharing,Kim_2021_CVPR,roth2020revisiting}. 
%%%

While sample-based objectives similarly suffer from the non-bijectivity of $s(\bullet, \bullet)$, the usage of sample-to-sample operations explicitly introduces intersample relational context \cite{roth2020revisiting} that allows for better structuring of samples within a class. 
However, just incorporating a sample-based contrastive operation into the training process is not a sufficient remedy as it re-introduces the sampling complexity issue; whose removal was what made proxy-based methods and their fast convergence attractive in the first place.

\subsection{Non-isotropy Regularization}
\label{subsec:nir}
%%% 
\textbf{Motivation.} To address the non-learnability of intraclass context in proxy-based DML, we therefore have to address the inherent issue of local isotropy in the learned sample-proxy distribution $p(\psi|\rho)$. However, to retain the convergence (and generalization) benefits of proxy-based methods, this has to be achieved without resorting to additional augmentations that move the overall objective away from a purely proxy-based one.
%%%%%
As such, we aim to find $p(\psi|\rho)$ whose optimization better resolves the distribution of sample representations $\psi$ around our proxy $\rho$.
This can be achieved by breaking down the fundamental issue of non-bijectivity in the used similarity measure $s(\psi, \rho)$, which (on its own) introduces non-unique sample-proxy relations.
%%%%%
To do so, we look for some regularization that specifically encourages unique sample-proxy relations to exists.
For such \textbf{unique} sample-proxy relations to exist, we must have access to some \textbf{bijective} and thus \textbf{invertible} (deterministic) translation $\psi = \tau(\zeta|\rho)$ which, given some residual $\zeta$ from some prior distribution $q(\zeta)$, allows to uniquely translate from the respective proxy $\rho$ to $\psi$. 
Given such a unique translation of samples and proxies within a class, the local alignment of samples would then no longer rely on relations to proxies and samples from different classes which, as noted, do not scale well locally and as training progresses.
%%%%%
Assuming proxies and sample representation to have the same dimensionality, this can be achieved through some affine transformation. However, to capture non-linear relations and proxy-to-sample translations, it is much more beneficial for $\tau$ to be non-linear. 

\textbf{Normalizing Flows.} Such invertible, non-linear functions are naturally expressed through Normalizing Flows (NF) or more generally invertible neural networks \cite{normflows,realnvp,glow,cinn,cinn_translate,cinn_imtoim,samesame}.
A Normalizing Flow can be generally seen as a transformation between two probability distributions, most commonly between simple, well-defined ones and complex multimodal ones \cite{nf_1,realnvp,nf_2,glow}.
%%%%%%
More specifically, we leverage flows similar to the one proposed in \cite{realnvp} and \cite{glow} (and as used e.g. in \cite{cinn,cinn_translate,cinn_imtoim,samesame,cinn_video}), which introduces a sequence of non-linear, but still invertible coupling operations as showcased in Fig. \ref{fig:setup} ($\leftrightharpoons$). Given some input representation $\psi$, a coupling block splits $\psi$ into $\psi_1$ and $\psi_2$, which are scaled and translated in succession with non-linear scaling and translation networks $\eta_1$ and $\eta_2$, respectively. Note that following \cite{glow}, each network $\eta_i$ provides both scaling $\eta_i^s$ and translation values $\eta_i^t$, such that 
\begin{equation}
\label{eq:coupling}
\begin{split}
    \psi^*_2 &= \psi_2 \odot \exp\left( \eta_1^s(\psi_1)  \right) + \eta^t_1(\psi_1)\\
    \psi^*_1 &= \psi_1 \odot \exp\left( \eta_2^s(\psi^*_2)  \right) + \eta^t_2(\psi^*_2)\\
    \psi^* &= [\psi_1^*, \psi_2^*]
\end{split}    
\end{equation}
where $\psi^*$ denotes $\psi$ after passing through a respective coupling block.
Successive application of different $\eta_i$ then gives our non-linear invertible transformation $\tau$ from some prior distribution over residuals $q(\zeta)$ with explicit density and CDF (for sampling) to our target distribution.

\textbf{Enforcing non-isotropy.} Consequently, our bijective $\tau$ (conditioned on proxies $\rho$) induces a new sample representation distribution $p(\tau(\zeta|\rho)|\rho)$ as \textit{pushforward} from our prior distribution of residuals $q(\zeta)$ which accounts for unique sample-to-proxy relations, and which we wish to impose over our learned sample distribution $p(\psi|\rho)$. This introduces our $\mathbb{N}$on-$\mathbb{I}$sotropy $\mathbb{R}$egularization ($\mathbb{NIR}$). 
% While we don't necessarily have access to an explicit formulation for maximum likelihood estimation for that, we can leverage the invertibility of $\tau$ to tackle this.
$\mathbb{NIR}$ can be naturally approached through maximization of the expected log-likelihood $\mathbb{E}_{x, \rho_{y_x}}\left[\log p\left(\psi(x)|\rho_{y_x}\right)\right]$ over sample-proxy pairs $(x, \rho_{y_x})$ similar to Eq. \ref{eq:pncapp}, but under the constraint that each distribution of samples around a respective proxy, $p(\psi|\rho)$, is a \textit{pushforward} of $\tau$ from our residual distribution $q(\zeta)$. This gives (see e.g. \cite{kobyzev2019normalizing})
\begin{equation}
\begin{split}
    \mathcal{L}_\mathbb{NIR} &= -\mathbb{E}_{x, \rho_{y_x}}[\log q\left(\tau^{-1}(\psi(x)|\rho_{y_x})\right)\\ &+ \log|\det J_{\tau^{-1}}(\tau^{-1}(\psi(x)|\rho_{y_x})|\rho_{y_x})|]
\end{split}
\end{equation}
with Jacobian $J$ for translation $\tau^{-1}$ and proxies $\rho_{y_x}$, where $y_x$ denotes the class of sample $x$. To arrive at above equation, we simply leveraged the change of variables formula
\begin{equation}
    p(\psi|\rho) = q(\tau^{-1}(\psi|\rho))|\det J_{\tau^{-1}}(\tau^{-1}(\psi|\rho)|\rho)|.
\end{equation}
In practice, by setting our prior $q(\zeta)$ to be a standard zero-mean unit-variance normal distribution $\mathcal{N}(0, 1)$, we get
\begin{equation}
\label{eq:nir}
\begin{split}
\mathcal{L}_\mathbb{NIR} &= \frac{1}{|\mathcal{B}|}\sum_{(x, \rho_{y_x})\sim\mathcal{B}}\left\Vert \tau^{-1}(\psi(x)|\rho_{y_x})\right\Vert^2_2\\ &- \log |\det J_{\tau^{-1}}(\tau^{-1}(\psi(x)|\rho_{y_x})|\rho_{y_x})|
\end{split}
\end{equation}
i.e. given sample representations $\psi(x)$, we project them onto our residual space $\zeta$ via $\tau^{-1}$ and compute Eq. \ref{eq:nir}. By selecting suitable normalizing flows such as GLOW \cite{glow}, we make sure that the Jacobian is cheap to compute.

\begin{table*}[t]
    \caption{\textit{Relative comparison.} We follow protocols proposed in \cite{roth2020revisiting}\protect\footnotemark, with no learning rate scheduling, to ensure exact comparability. The results show significant improvements over very strong proxy objectives on all benchmarks, but especially on CUB200 and CARS196 where a more significant number of samples per class is available.}
    \vspace{-3pt}
 \footnotesize
  \setlength\tabcolsep{1.4pt}
  \centering
  %\begin{tabular}{l|c|ccc|c}
  \resizebox{0.95\textwidth}{!}{
  \begin{tabular}{l || c | c | c || c | c| c || c | c | c}
     \toprule
     \multicolumn{1}{l}{\textsc{Benchmarks}$\rightarrow$} & \multicolumn{3}{c}{\textsc{CUB200-2011}} & \multicolumn{3}{c}{\textsc{CARS196}} & \multicolumn{3}{c}{\textsc{SOP}} \\
     \midrule
     \textsc{Approaches} $\downarrow$ & R@1 & NMI & mAP@1000 & R@1 & NMI & mAP@1000 & R@1 & NMI & mAP@1000\\
    \midrule
    % \rowcolor{vvlightgray} 
    \textbf{Multisimilarity} & 62.8 $\pm$ 0.2 & 67.8 $\pm$ 0.4 & 31.1 $\pm$ 0.3 & 81.6 $\pm$ 0.3 & 69.6 $\pm$ 0.5 & 31.7 $\pm$ 0.1 & 76.0 $\pm$ 0.1  & 89.4 $\pm$ 0.1  & 43.3 $\pm$ 0.1 \\
    \hline
    % \rowcolor{vvlightgray} 
    \textbf{ProxyAnchor} \cite{kim2020proxy} & 64.4 $\pm$ 0.3 & 68.4 $\pm$ 0.2 & 33.2 $\pm$ 0.3 & 82.4 $\pm$ 0.4 & 69.0 $\pm$ 0.3 & 34.2 $\pm$ 0.3 & 78.0 $\pm$ 0.1 & 90.1 $\pm$ 0.1 & 45.5 $\pm$ 0.1\\    
    + $\mathbb{NIR}$ & 66.0 $\pm$ 0.3 & 69.6 $\pm$ 0.1 & 34.2 $\pm$ 0.2& 85.2 $\pm$ 0.3 & 71.6 $\pm$ 0.3 & 36.4 $\pm$ 0.2 & 78.9 $\pm$ 0.1 & 90.4 $\pm$ 0.1 & 46.5 $\pm$ 0.1\\    
    \hline
    % \rowcolor{vvlightgray} 
    \textbf{ProxyNCA} \cite{proxynca} & 64.2 $\pm$ 0.2 & 68.6 $\pm$ 0.3 & 33.1 $\pm$ 0.2 & 82.1 $\pm$ 0.4 & 68.2 $\pm$ 0.2 & 32.4 $\pm$ 0.5 & 78.3 $\pm$ 0.1 & 90.0 $\pm$ 0.1 & 45.5 $\pm$ 0.1\\    
    + $\mathbb{NIR}$ & 66.1 $\pm$ 0.2 & 69.8 $\pm$ 0.2 & 34.3 $\pm$ 0.1 & 84.3 $\pm$ 0.3 & 70.6 $\pm$ 0.6 & 34.5 $\pm$ 0.3 & 79.1 $\pm$ 0.1 & 90.2 $\pm$ 0.1 & 46.2 $\pm$ 0.1 \\
    \hline
    % \rowcolor{vvlightgray} 
    \textbf{SoftTriplet} \cite{softriple} & 62.3 $\pm$ 0.3 & 68.2 $\pm$ 0.2 & 31.6 $\pm$ 0.2 & 80.7 $\pm$ 0.2 & 66.4 $\pm$ 0.3 & 30.4 $\pm$ 0.2 & 76.9 $\pm$ 0.2 & 89.6 $\pm$ 0.1 & 43.5 $\pm$ 0.1 \\    
    + $\mathbb{NIR}$ & 63.8 $\pm$ 0.4 & 68.5 $\pm$ 0.2 & 34.0 $\pm$ 0.4 & 83.4 $\pm$ 0.4 & 68.8 $\pm$ 0.5 & 35.5 $\pm$ 0.2 & 77.6 $\pm$ 0.1 & 90.0 $\pm$ 0.1 & 44.9 $\pm$ 0.1\\        
    \bottomrule 

    \end{tabular}}
    \label{tab:relative_results}
    \vspace{-3pt}
 \end{table*}

\begin{table*}[t]
    \caption{\textit{Literature comparison} using ProxyAnchor (PA) + $\mathbb{NIR}$. Across backbones/dim. and benchmarks, we find competitive and even state-of-the-art performance against much more complex methods. \red{$^X$}: Combination of pooling operations in backbone as done in \cite{kim2020proxy}. \textbf{Bold} denotes best in a given \textit{Architecture/Dimensionality} setting. \blue{\textbf{Bluebold}} denotes best overall.}
    \vspace{-3pt}
    \setlength\tabcolsep{1.5pt}
    \footnotesize
    \centering

\resizebox{0.95\textwidth}{!}{
\begin{tabular}{l | l | l || c | c | c || c | c | c || c | c | c}
     \toprule
     \multicolumn{3}{l}{\textsc{Benchmarks} $\rightarrow$} & \multicolumn{3}{c}{\textsc{CUB200} \cite{cub200-2011}} & \multicolumn{3}{c}{\textsc{CARS196} \cite{cars196}} & \multicolumn{3}{c}{\textsc{SOP} \cite{lifted}}\\
     \midrule
     \textsc{Methods} $\downarrow$ & Venue & Arch/Dim. & R@1 & R@2 & NMI & R@1 & R@2 & NMI & R@1 & R@10 & NMI\\
     \midrule
     \hline
    %  Margin \cite{margin} & \textit{ICCV '17} &R50/128 & 63.6 & 74.4 & 69.0 & 79.6 & 86.5 & 69.1 & 72.7 & 86.2 & 90.7\\
     Div\&Conq \cite{Sanakoyeu_2019_CVPR} & \textit{CVPR '19} & R50/128 & 65.9 & 76.6 & 69.6 & 84.6 & 90.7 & 70.3 & 75.9 & 88.4 & 90.2\\
     MIC \cite{mic} & \textit{ICCV '19} & R50/128 & 66.1 & 76.8 & 69.7 & 82.6 & 89.1 & 68.4 & 77.2 & 89.4 & 90.0\\
     PADS \cite{roth2020pads} & \textit{CVPR '20} &R50/128 & 67.3 & 78.0 & 69.9 & 83.5 & 89.7 & 68.8 & 76.5 & 89.0 & 89.9\\
     RankMI \cite{rankmi} & \textit{CVPR '20} & R50/128 & 66.7 & 77.2 & \textbf{71.3} & 83.3 & 89.8 & 69.4 & 74.3 & 87.9 & 90.5 \\
     \hline
    \multirow{2}{*}{PA+$\mathbb{NIR}$} & - & R50/128 & 66.9 $\pm$ 0.2 & 77.7 $\pm$ 0.3 & 69.8 $\pm$ 0.2 & 85.3 $\pm$ 0.2 & 91.1 $\pm$ 0.2 & 72.1 $\pm$ 0.2 & \textbf{79.6 $\pm$ 0.1} & \textbf{90.7 $\pm$ 0.1} & \textbf{90.5 $\pm$ 0.1}\\
     & - & R50/128\red{$^X$} & \textbf{67.9 $\pm$ 0.2} & \textbf{78.3 $\pm$ 0.2} & \textbf{71.4 $\pm$ 0.4} & \textbf{86.5 $\pm$ 0.3} & \textbf{92.0 $\pm$ 0.2} & \textbf{72.7 $\pm$ 0.2} & 79.4 $\pm$ 0.1 & \textbf{90.7 $\pm$ 0.1} & \textbf{90.6 $\pm$ 0.1} \\     
    \hline
    %%%%%%%%%%%%%%%%%%%%%%%%%%%%%%%%%%%%%%%%%%%%%%%%%%%%%
    %%%%%%%%%%%%%%%%%%%%%%%%%%%%%%%%%%%%%%%%%%%%%%%%%%%%%    
    %  Group \cite{elezi2020grouploss} & \textit{ECCV '20} &IBN/512 & 65.5 & 77.0 & 69.0 & 85.6 & 91.2 & 72.7 & 75.1 & 87.5 & \blue{\textbf{90.8}}   \\    
    %  Multisimilarity\red{$^{c}$} \cite{multisimilarity} & IBN/512 & 65.7 & 77.0 & - & 84.1 & 90.4 & -  & 78.2 & 90.5 & -   \\
    %  DR-MS \cite{dutta2020orthogonalunsupdml} & \textit{TAI '20}& IBN/512 & 66.1 & 77.0 & - & 85.0 & 90.5 & -  & - & - & -   \\
     ProxyAnchor (PA) \cite{kim2020proxy} & \textit{CVPR '20}& IBN/512 \red{$^X$}& 68.4 & 79.2 & - & 86.8 & 91.6 & -  & 79.1 & 90.8 & -   \\ 
     ProxyGML \cite{Zhu2020graphdml} & \textit{NeurIPS '20} & IBN/512 & 66.6 & 77.6 & 69.8 & 85.5 & 91.8 & 72.4  & 78.0 & 90.6 & 90.2   \\     
     DRML \cite{drml} & \textit{ICCV '21} & IBN/512 & 68.7 & 78.6 & 69.3 & 86.9 & 92.1 & 72.1 & 71.5 & 85.2 & 88.1 \\
     PA + \textit{MemVir} \cite{memvir} & \textit{ICCV '21} & IBN/512 & 69.0 & 79.2 & - & 86.7 & 92.0 & - & \textbf{79.7} & \textbf{91.0} & - \\
     \hline
    \multirow{2}{*}{PA+$\mathbb{NIR}$} & - & IBN/512 & 69.4 $\pm$ 0.2 & 79.7 $\pm$ 0.2 & \textbf{71.1 $\pm$ 0.1} & 87.1 $\pm$ 0.2 & 92.5 $\pm$ 0.1 & 73.1 $\pm$ 0.2 & 79.4 $\pm$ 0.1 & 90.5 $\pm$ 0.1 & 90.3 $\pm$ 0.2 \\     
    & - & IBN/512\red{$^X$}  & \textbf{70.1 $\pm$ 0.1} & \textbf{80.1 $\pm$ 0.2} & \textbf{71.0 $\pm$ 0.2} & \textbf{87.9 $\pm$ 0.2} & \textbf{92.8 $\pm$ 0.1} & \textbf{73.7 $\pm$ 0.2} & 79.3 $\pm$ 0.1 & 90.4 $\pm$ 0.1 & 90.2 $\pm$ 0.2 \\     
    \hline
    %%%%%%%%%%%%%%%%%%%%%%%%%%%%%%%%%%%%%%%%%%%%%%%%%%%%%
    %%%%%%%%%%%%%%%%%%%%%%%%%%%%%%%%%%%%%%%%%%%%%%%%%%%%%
    %  S2SD \cite{s2sd} & R50/128 & 68.9 $\pm$ 0.3 & 79.0 $\pm$ 0.3 & 72.1 $\pm$ 0.4 & 87.6 $\pm$ 0.2 & 92.7 $\pm$ 0.2 & 72.3 $\pm$ 0.2& 80.2 $\pm$ 0.2 & 91.5 $\pm$ 0.1 & 90.9 $\pm$ 0.1\\     
    %  NormSoft \cite{zhai2018classification} & \textit{BMVC '19} & R50/512 & 61.3 & 73.9 &  -   & 84.2 & 90.4 &  -   & 78.2 & 90.6 &  -    \\    
     EPSHN \cite{epshn} & \textit{WACV '20} & R50/512 & 64.9 & 75.3 &  -   & 82.7 & 89.3 &  -  & 78.3 & 90.7 &  -\\
     Circle \cite{circle} & \textit{CVPR '20} & R50/512 & 66.7 & 77.2 & - & 83.4 & 89.7 & - & 78.3 & 90.5 & - \\
     DiVA \cite{milbich2020diva} & \textit{ECCV '20} & R50/512 & 69.2 & 79.3 & 71.4 & 87.6 & 92.9 & 72.2 & 79.6 & 91.2 & 90.6 \\
     DCML-MDW \cite{Zheng_2021_CVPR_compositional} & \textit{CVPR '21} & R50/512 & 68.4 & 77.9 & 71.8 & 85.2 & 91.8 & 73.9 & 79.8 & 90.8 & \blue{\textbf{90.8}} \\
     \hline
    \multirow{2}{*}{PA+$\mathbb{NIR}$} & - & R50/512 & 69.1 $\pm$ 0.2 & 79.6 $\pm$ 0.2 & 72.0 $\pm$ 0.2 & 87.7 $\pm$ 0.2 & 92.5 $\pm$ 0.1 & 74.2 $\pm$ 0.2 & \blue{\textbf{80.7 $\pm$ 0.1}} & \blue{\textbf{91.5 $\pm$ 0.1}} & \blue{\textbf{90.9 $\pm$ 0.1}} \\
    & - & R50/512\red{$^X$} & \blue{\textbf{70.5 $\pm$ 0.1}} & \blue{\textbf{80.6 $\pm$ 0.2}} & \blue{\textbf{72.5 $\pm$ 0.3}} & \blue{\textbf{89.1 $\pm$ 0.2}} & \blue{\textbf{93.4 $\pm$ 0.2}} & \blue{\textbf{75.0 $\pm$ 0.3}}  & 80.4 $\pm$ 0.1 & 91.4 $\pm$ 0.2 & 90.6 $\pm$ 0.1 \\      
     \bottomrule
\end{tabular}}

\vspace{-3pt}
\label{tab:sota}
\end{table*}

\textbf{$\mathbb{NIR}$ for proxy-based DML.} As $\mathbb{NIR}$ targets the alignment of samples around proxies, we still need to learn the global alignment of proxies through proxy-based DML $\mathcal{L}_\text{PDML}(\psi, \mathcal{P})$. This gives the full training objective
\begin{equation}
    \mathcal{L} = f(\mathcal{L}_\mathbb{NIR}) + \omega \cdot \mathcal{L}_\text{PDML}(\psi, \mathcal{P})
\end{equation}
where $f(\cdot)$ is a monotonous function of $\mathcal{L}_\mathbb{NIR}$ to match the scaling and training dynamics of $\mathcal{L}_\text{PDML}$ without changing the invertibility constraint. As most proxy-based $\mathcal{L}_\text{PDML}$ utilize exponential components, we simply use $f(\cdot) = \exp(\cdot)$. 
Full optimization over $\mathcal{L}$ then learns proxies while uniquely resolving sample placement around them.
%%%%%%
More specifically, backpropagating through $\mathcal{L}_\mathbb{NIR}$ optimizes for sample alignment around proxies $\rho$, the translation $\tau$ and provides updates to the proxies, though we found the latter to not be a necessity, as proxies primarily serve to resolve global alignment of sample clusters. 

$\mathbb{NIR}$-proxy-DML has several advantages. Firstly, the final sample-proxy distribution optimized for directly addresses issues of local isotropy to better resolve local intraclass structure, as the retention of a unique sample distribution around each proxies requires implicit knowledge about the intraclass alignment of each class sample around their respective proxies. Secondly, unlike ProxyNCA-like objectives (see the previous section), we do not assume the same concentration of samples per class as assumed in e.g. Eq. \ref{eq:pnca}. Instead, the non-linear translation conditioned on the class proxy can introduce class-dependent concentrations when needed. Finally, being able to directly resolve local structures can potentially benefit convergence of these methods.

%%%%%%%%%%%%%%%%%%%%%%%%%%%%%%%%%%%%%%%%%%%%%%%%%%%%%%%%%%%%%%%%%%%%%%%%
%%%%%%%%%%%%%%%%%%%%%%%%%%%%%%%%%%%%%%%%%%%%%%%%%%%%%%%%%%%%%%%%%%%%%%%%
\section{Experiments}
\label{sec:experiments}
This section lists experimental details (\S\ref{subsec:exp_details}), showcases significant benefits of $\mathbb{NIR}$ for proxy-based DML (\S\ref{subsec:exp_nir}) and highlights the impact on convergence and training times in \S\ref{subsec:exp_convergence}. We also study quantitative impacts on learned representation spaces (\S\ref{subsec:exp_structure}), provide method ablations in \S\ref{subsec:exp_ablations} and investigate self-regulatory properties of $\mathbb{NIR}$ (\S\ref{subsec:exp_selfreg}).

\subsection{Experimental Details}
\label{subsec:exp_details}
\textbf{Implementations} use \texttt{PyTorch} \cite{pytorch}. ImageNet\cite{imagenet}-pretrainings are taken from \texttt{torchvision}\cite{torchvision} and \texttt{timm}\cite{timm}. 
% Following exisiting protocols \cite{margin,mic,milbich2020diva,kim2020proxy,roth2020revisiting,musgrave2020metric}, we freeze all Batch-Normalization layers during training. 
Our experiments were run on compute servers with NVIDIA 2080Ti. Our Normalizing Flow utilizes 8 coupling blocks and subnets $\eta$ comprising linear layers with 128 nodes. Optimization is done using Adam \cite{adam} (learning rate $10^{-5}$, weight decay $4\cdot 10^{-3}$, \cite{roth2020revisiting}). We set $\omega\in [0.001, 0.01]$ depending on the choice for $\mathcal{L}_\text{PDML}$. In general, we found consistent improvements in this interval. Following \cite{proxynca,proxyncapp,kim2020proxy} we utilize a high learning rate multiplication for the proxies (4000). We also saw this helping the Normalizing Flow and use 50 for all experiments. Finally, we found a warmup epoch to help; adapting the translation $\tau$ over pretrained features first before joint training. 

\textbf{Datasets.} We use the standard benchmarks CUB200-2011\cite{cub200-2011} (11,788 bird images, 200 classes), CARS196\cite{cars196} (16,185 car images, 196 classes) and Stanford Online Products\cite{lifted} (SOP, 120,053 images, 22,634 product instances).

\subsection{Effectiveness of Non-isotropy Regularisation}
\label{subsec:exp_nir}
To evaluate the relative benefits of $\mathbb{NIR}$,
% \Zeynep{should we move the rest to the imlementation details?} 
we follow protocols proposed in \cite{roth2020revisiting} to encourage exact comparability with no learning rate scheduling. Initial tuning of newly introduced hyperparameters is done on a random $15\%$ validation split (see e.g. \cite{kim2020proxy,roth2020revisiting}). 
% \Zeynep{until here} 
Note that we don't perform joint hyperparameter tuning of auxiliary proxy objectives $\mathcal{L}_\text{DML}$ and $\mathbb{NIR}$ to see how well $\mathbb{NIR}$ can be applied to blackbox proxy approaches. For $\mathcal{L}_\text{DML}$, we select ProxyAnchor \cite{kim2020proxy}, SoftTriplet \cite{softriple}(10 centroids for CUB200-2011/CARS196 and 2 for SOP) and ProxyNCA \cite{proxynca} following Eq. \ref{eq:pnca_ada}. Results over multiple seeds are provided in Table \ref{tab:relative_results}, showing that generalization significantly improves for all proxy objectives across metrics and benchmarks, even for objectives with more than one proxy per class s.a. SoftTriple. 
For the latter, we find $\mathbb{NIR}$ to also improve convergence properties as shown in \S\ref{subsec:exp_convergence}. 
Especially for datasets where a reasonable amount of samples per class is available (s.a. CUB200-2011 \& CARS196) to learn meaningful class distributions we see major improvements, e.g. for state-of-the-art ProxyAnchor from $82.4\%$ to $85.2$ \& $64.4$ to $66.0\%$ on CARS196 \& CUB200-2011, respectively. However, even for datasets such as SOP with a small number of samples per class a consistent performance improvement can be seen, highlighting the general benefit of $\mathbb{NIR}$ for proxy-based DML. 
% \Zeynep{Why both Proxy anchor and proxy NCA both increase by 2\% on CUB with NIR whereas for SotTriplet it is not the same? on Cars this is not the same though: softtriplet also increases significantly. Is there a reason?}

To compare to the overall corpus of DML methods, we also provide a literature comparison in Table \ref{tab:sota}, with approaches divided based on backbone architecture and embedding dimensionality; both of which drive generalization performance independent of DML objectives \cite{roth2020revisiting,musgrave2020metric}. Results reported here use stepwise learning rate scheduling at most twice, with parameters determined by performance on a random, $15\%$ validation subset \cite{kim2020proxy,roth2020revisiting,milbich2020diva}. As can be seen, $\mathbb{NIR}$-equipped ProxyAnchor achieves competitive performance across settings and benchmarks with a new highest overall score. In addition, PA+$\mathbb{NIR}$ beats much more complex methods such as \textit{DiVA}\cite{milbich2020diva} using joint multitask and self-supervised training or MIC\cite{mic} using external feature mining. This supports the benefit of learning global alignments with proxies while jointly refining local sample alignment. Taking into account that $\mathbb{NIR}$ retains the superior convergence of proxy-based approaches (see \S\ref{subsec:exp_convergence}), this makes $\mathbb{NIR}$ very attractive for practical usage, and provides a strong proof-of-concept on the benefits of intraclass context for proxy-DML.
% \Zeynep{I think we should expand this explanation a bit and talk about the results separately for different datasets. I would also try to argue why this is happening.}

\subsection{Convergence Properties}
\label{subsec:exp_convergence}
\begin{figure}[t!]
    \centering
    \includegraphics[width=0.47\textwidth]{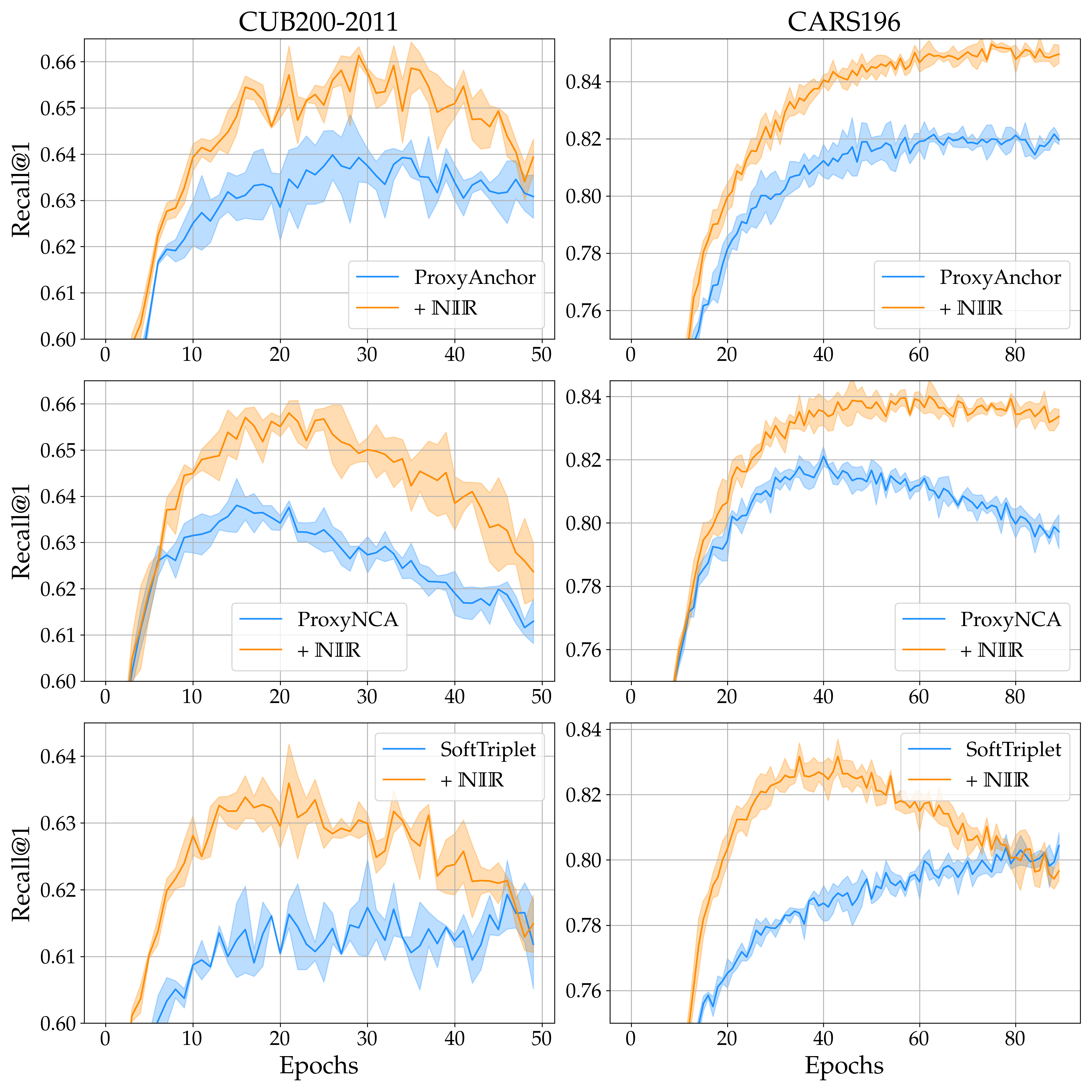}
    \vspace{-5pt}
    \caption{\textit{Impact of} $\mathbb{NIR}$ \textit{on convergence.} We find that $\mathbb{NIR}$ increases generalisation performance while retaining or even improving the fast convergence behaviour.}
    \label{fig:convergence}
    \vspace{-5pt}
\end{figure}

A primary motivation for $\mathbb{NIR}$, besides generalization improvements, is to retain fast convergence speeds. We investigate this following the same setup used for Tab. \ref{tab:relative_results} (\S\ref{subsec:exp_nir}). Results visualized in Fig. \ref{fig:convergence} show mean test generalisation performance after every epoch \cite{Sanakoyeu_2019_CVPR,proxynca,kim2020proxy} (unlike Tab. \ref{tab:relative_results} showing overall mean test performance) for every auxiliary proxy objective. Besides significant improvements in generalisation performance, convergence speeds and behaviour are either retained or even improved e.g. for SoftTriple, presumably due to better resolved class structures allowing for better alignment of multiple learned class centroids. Furthermore, the addition of the Normalizing Flow adds only limited additional computational overhead since we operate in feature space. Especially with respect to the large backbone, we find changes in walltime and required additional GPU memory to be negligible ($< 1\%$). 
% \Zeynep{I would keep one of these methods (whichever is the most recent) as the different rows show the same behavior (we could move the other two methods to the supplementary) and explain the plots more.} \Zeynep{the softTriplet results converge to the same point at the end of the epochs. when you report results in Table 1, which epoch do you use? Are the results in table 1 and this section match? The SOP results seem to be different.}
% This makes $\mathbb{NIR}$ extremely attractive for use in practice alongside commonly employed proxy-based DML.

\begin{table}[t!]
\caption{\textit{Change in Structural Properties.} Applying $\mathbb{NIR}$ increase feature diversity $\rho$ and uniformity of learned representations $G_2$, reduces overclustering $\pi_\text{density}$ and encourages a higher degree in class concentration difference between classes ($\sigma_\kappa^2$).}
\vspace{-5pt}
\footnotesize
\setlength\tabcolsep{1.4pt}
\centering
%\begin{tabular}{l|c|ccc|c}
\resizebox{0.47\textwidth}{!}{
    \begin{tabular}{l | l || c | c | c | c }
    \toprule
    \textbf{Dataset} & \textbf{Setup} & $\rho \downarrow$ & $\pi_\text{density} \uparrow$ & $\sigma^2_\kappa \uparrow$ & $G_2 \downarrow$ \\
    \midrule
    \multirow{2}{*}{CUB200} & PA & 0.19 $\pm$ 0.01 & 0.68 $\pm$ 0.04 & 0.37 $\pm$ 0.02 & 0.078 $\pm$ 0.001\\
    & + $\mathbb{NIR}$& \textbf{0.13} $\pm$ 0.02 & \textbf{0.79} $\pm$ 0.03 & \textbf{0.44} $\pm$ 0.02 & \textbf{0.072} $\pm$ 0.002 \\
    \hline
    \multirow{2}{*}{CARS196} & PA & 0.17 $\pm$ 0.01 & 0.59 $\pm$ 0.01 & 0.32 $\pm$ 0.01 & 0.079 $\pm$ 0.001 \\
    & + $\mathbb{NIR}$ & \textbf{0.13} $\pm$ 0.01 & \textbf{0.68} $\pm$ 0.02 & \textbf{0.38} $\pm$ 0.02 & \textbf{0.072} $\pm$ 0.001 \\
    \bottomrule
    \end{tabular}
}
\label{tab:structural_properties}
\vspace{-5pt}
\end{table}

\begin{figure}[t!]
    \centering
    \includegraphics[width=0.47\textwidth]{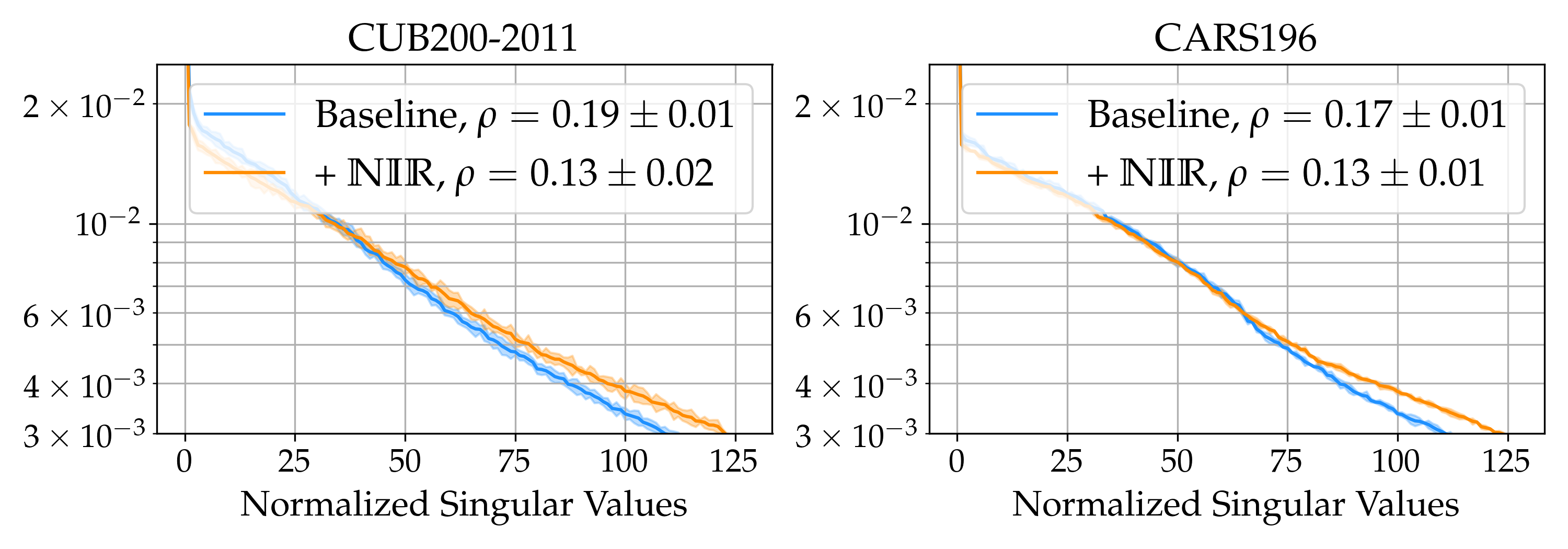}
    \vspace{-5pt}
    \caption{\textit{Singular Value Distribution} used to estimate feature diversity $\rho$, showing $\mathbb{NIR}$ to introduce more directions of variance.}
    \label{fig:feat_diversity}
    \vspace{-5pt}
\end{figure}

\subsection{Qualitative differences in alignment}
\label{subsec:exp_structure}
In this section, we investigate how $\mathbb{NIR}$ changes the structural properties of the representation spaces learned by proxy-based objectives. For our experiments, we select ProxyAnchor as stand-in for $\mathcal{L}_\text{PDML}$. We then compare the structure of representation space learned with and without $\mathbb{NIR}$ by looking at different structural metrics: 
%%%%%%%
\textbf{(1)} Feature richness measured by spectral decay \cite{roth2020revisiting} $\rho(\Psi) = \text{KL}(\mathcal{U}||S(\Psi))$ with singular value decomposition $S(\Psi)$ of feature space $\Psi$. $\rho(\Psi)$ measures the number of significant directions of variance in the learned feature space - lower scores indicate a higher feature variety, linked to improved generalisation in \cite{roth2020revisiting,milbich2020diva}.
% , who explicitly optimize for this property by joint multitask and self-supervised learning. 
%%%%%%%
\textbf{(2)} Representational Uniformity/Density \cite{roth2020revisiting} $\pi_\text{density} = \nicefrac{\pi_\text{intra}}{\pi_\text{inter}}$, which measures the ratio of mean intraclass and interclass distance $\pi_\text{intra}$ and $\pi_\text{inter}$, respectively. $\pi_\text{density}$ relates class concentration against overall alignment across the hypersphere. Higher values indicate lower class concentrations and overclustering, a potential link to better generalisation \cite{roth2020revisiting}. 
%%%%%%%
\textbf{(3)} Embedding space uniformity \cite{wang2020understanding} $G_{t=2}(u, v) = e^{-t\left\Vert u - v\right\Vert^2_2}$ evaluating uniformity of embedding spaces measured by a Radial Basis Function kernel \cite{rbf}. Lower values have been linked in \cite{wang2020understanding} to improved downstream performance in contrastive self-supervised learning. 
%%%%%%%
\textbf{(4)} Variance of class-concentrations $\sigma_\kappa^2$ approximated by the mean distances to the class centers of mass (relative to the mean interclass distance $\pi_\text{inter}$ to account for the overall scale of representations). As $\mathbb{NIR}$ allows for different class-conditional distributions to be learned, we assume a higher $\sigma_\kappa^2$.

% \Zeynep{the build up to the experiments is more detailed than the experimental results. I would write the previous four settings as separate paragraphs and quote results and discuss about these results separately.
% It is not clear from the description below which of these settings fair in which way.}
Results in  Tab. \ref{tab:structural_properties} indeed show higher feature diversity $\rho$ (over $30\%$, cf. Fig. \ref{fig:feat_diversity} for the sorted singular value spectra), reduced overclustering ($>15\%$) as measured by $\pi_\text{density}$ and increased uniformity in learned representation spaces (up to $9\%$ as evaluated via $G_2$) - all linked to better generalization as noted above.
% \footnote{As done e.g. in \textit{DiVA}\cite{milbich2020diva} with additional training tasks and self-supervision. However, in our case, with negligible overhead and convergence drawbacks.}. 
This links well with our initial motivation for $\mathbb{NIR}$ to allow for better explicit resolution of local structures and clusters, which requires local separability of representations and the introduction of auxiliary features \cite{roth2020revisiting,milbich2020diva}. We also find that the variance of class-concentrations increases, supporting that $\mathbb{NIR}$ helps proxy-based objectives learn class-dependent sample distributions.

\begin{table}[t]
    \caption{\textit{Structural Ablations.} }
    \vspace{-5pt}
 \footnotesize
  \setlength\tabcolsep{1.4pt}
  \centering
  %\begin{tabular}{l|c|ccc|c}
  \resizebox{0.47\textwidth}{!}{
  \begin{tabular}{l || c | c || c | c }
     \toprule
     \multicolumn{1}{l}{\textsc{Benchmarks}$\rightarrow$} & \multicolumn{2}{c}{\textsc{CUB200-2011}} & \multicolumn{2}{c}{\textsc{CARS196}} \\
     \midrule
     \multirow{2}{*}{\textsc{Setups} $\downarrow$} & \multirow{2}{*}{R@1} & mAP & \multirow{2}{*}{R@1} & mAP \\
     & & @1000 & & @1000 \\
    \midrule
    % \rowcolor{vlightgray}    
    % Multisimilarity & 62.8 $\pm$ 0.2 & 31.1 $\pm$ 0.3 & 81.6 $\pm$ 0.3 & 31.7 $\pm$ 0.1\\    
    % \rowcolor{vlightgray}    
    % ProxyAnchor (PA) & 64.4 $\pm$ 0.3 & 33.2 $\pm$ 0.3 & 82.4 $\pm$ 0.4 & 34.2 $\pm$ 0.3\\
    % \rowcolor{vvlightgray}    
    PA + $\mathbb{NIR}$ & 66.0 $\pm$ 0.3 & 34.2 $\pm$ 0.2 & 85.2 $\pm$ 0.2 & 36.4 $\pm$ 0.2\\
    \midrule    
    % \multicolumn{5}{>{\columncolor[gray]{.9}}l}{\textbf{(a) Importance of auxiliary proxies}} \\    
    % \hline    
    % % Without ($\omega = 0$) & & & & \\
    % % With Prototypes ($\omega = 0$) & & & & \\    
    \multicolumn{5}{>{\columncolor[gray]{.9}}l}{\textbf{(a) Normalizing Flows Training}} \\        
    \hline
    $f(\cdot)$ = SoftPlus & 66.3 $\pm$ 0.2 & 34.0 $\pm$ 0.2 & 85.1 $\pm$ 0.2 & 36.3 $\pm$ 0.3 \\    
    $f(\cdot, t=0.3)$ = Exp & 66.1 $\pm$ 0.2 & 34.1 $\pm$ 0.1 & 85.0 $\pm$ 0.2 & 36.2 $\pm$ 0.2\\    
    $f(\cdot, t=3)$ = Exp & 65.8 $\pm$ 0.3 & 33.8 $\pm$ 0.2 & 84.8 $\pm$ 0.3 & 36.1 $\pm$ 0.2\\
    \hline
    Grad. Clipping & 65.9 $\pm$ 0.1 & 34.2 $\pm$ 0.1 & 85.3 $\pm$ 0.1 & 36.5 $\pm$ 0.1 \\
    No Proxy Backprop & 66.2 $\pm$ 0.4 & 34.1 $\pm$ 0.3 & 85.0 $\pm$ 0.2 & 36.3 $\pm$ 0.1\\ 
    w/ Negative Pairs & 64.9 $\pm$ 0.3 & 33.8 $\pm$ 0.3 & 83.5 $\pm$ 0.4 & 34.9 $\pm$ 0.3 \\ 
    $\omega = 0$ & 60.0 $\pm$ 0.4 & 30.1 $\pm$ 0.3 & 73.5 $\pm$ 0.6 & 27.2 $\pm$ 0.5 \\
    \hline
    \multicolumn{5}{>{\columncolor[gray]{.9}}l}{\textbf{(b) Normalizing Flows Architecture}} \\        
    \hline
    D15 - W64 & 66.5 $\pm$ 0.5 & 34.0 $\pm$ 0.2 & 85.1 $\pm$ 0.4 & 36.6 $\pm$ 0.3\\    
    D5 - W512 & 66.1 $\pm$ 0.3 & 34.1 $\pm$ 0.1 & 84.9 $\pm$ 0.2 & 36.1 $\pm$ 0.2 \\    
    D3 - W1024 & 65.8 $\pm$ 0.4 & 34.1 $\pm$ 0.2 & 85.3 $\pm$ 0.1 & 36.3 $\pm$ 0.2\\
    \hline
    Condition - Start & 66.4 $\pm$ 0.4 & 34.1 $\pm$ 0.2 & 85.1 $\pm$ 0.2 & 36.2 $\pm$ 0.3\\    
    Condition - Mid & 66.1 $\pm$ 0.2 & 34.1 $\pm$ 0.2 & 84.9 $\pm$ 0.2 & 36.0 $\pm$ 0.2\\
    Condition - End & 65.7 $\pm$ 0.2 & 33.8 $\pm$ 0.3 & 84.8 $\pm$ 0.3 & 36.0 $\pm$ 0.2\\
    % Condition - Dense + $\mathcal{T}$ & 66.3 $\pm$ 0.2 & 34.2 $\pm$ 0.1 & 84.8 $\pm$ 0.4 & 35.9 $\pm$ 0.3\\
    \bottomrule
    \end{tabular}}
    \label{tab:architecture_ablations}
    \vspace{-5pt}
 \end{table}

\subsection{Ablations}
\label{subsec:exp_ablations}
We now ablate $\mathbb{NIR}$. 
Results are provided in Tab. \ref{tab:architecture_ablations}. We use ProxyAnchor as stand-in for $\mathcal{L}_\text{PDML}$

\textbf{(a) Training Normalizing Flows.} We first ablate the scaling $f(\cdot)$ (\S\ref{subsec:nir}, Eq. \ref{eq:nir}). As can be seen, the exact choice of exponential function does not matter, with changes in temperature ($f(\cdot, t)$) or a Softplus ($f(x) = \log(1 + \exp x)$) performing similarly. We also experiment with gradient clipping (``\textit{Grad. Clipping}''), but found no benefits.
Furthermore, we investigate joint negative log-likelihood (NLL) \textit{maximization} for sample-proxy-pairs with different classes (``\textit{w/ Negative Pairs)''} following Eq. \ref{eq:nir}, but found no benefits over just minimizing the NLL for same-class pairs. This supports our hypothesis that the benefits of $\mathbb{NIR}$ indeed lie in improved resolution of class-local structures.
This is also supported through the minimal impact of proxy updates through $\mathcal{L}_\mathbb{NIR}$ (``\textit{No Proxy Backprop}''), as proxies primarily help with global cluster alignment while $\mathbb{NIR}$ is developed for local refinement. 
Completely removing $\mathcal{L}_\text{PDML}$
% \footnote{Here, we saw benefits from joint NLL maximization of negative pairs.} 
(``$\omega = 0$'') gives similar insights - while we see decent performance solely through non-isotropy regularisation, the absence of a global alignment objective for proxies incorporated through $\mathcal{L}_\text{PDML}$ (``$\omega = 0$'') results in a notable performance drop and reduction in convergence speeds. 

\textbf{(b) Normalizing Flows Architecture.} We ablate the number of coupling blocks (\textit{D)} and subnets widths (\textit{W}) used in $\mathbb{NIR}$ and find dataset-dependent optima with slight improvements over the default (\S\ref{subsec:exp_details}). For consistency, we report all other results using the default setup.
We also examine the conditioning of the Normalizing Flow - inserting proxies at the start, mid or end (``\textit{Condition start/mid/end}'', as opposed to every coupling block by default). As can be seen, the exact choice of conditioning influences generalization performance somewhat, but with significant improvements regardless of the exact conditioning. 
% We also find additional transformations $\mathcal{T}$ (via MLP, ``\textit{Condition - Dense +}$\mathcal{T}$'') on top of the condition to offer no major benefits.
% \Zeynep{which results in the table indicate this?}.
Overall, these results show that in an applied setting, further improvements can be found by more aggressive, but not necessarily principled hyperparameter tuning of $\mathbb{NIR}$. 

% \Zeynep{in general, you could walk through the table a bit, e.g. quote some interesting results and argue why this is happening.}

\begin{table}[t]
    \caption{\textit{Self-Regularisation} by reversing $\tau$ to generate synthetic samples does not benefit generalisation.}
\vspace{-5pt}
 \footnotesize
  \setlength\tabcolsep{1.4pt}
  \centering
  %\begin{tabular}{l|c|ccc|c}
  \resizebox{0.47\textwidth}{!}{
  \begin{tabular}{l || c | c || c | c }
     \toprule
     \multicolumn{1}{l}{\textsc{Benchmarks}$\rightarrow$} & \multicolumn{2}{c}{\textsc{CUB200-2011}} & \multicolumn{2}{c}{\textsc{CARS196}} \\
     \midrule
     \multirow{2}{*}{\textsc{Approach} $\downarrow$} & \multirow{2}{*}{R@1} & mAP & \multirow{2}{*}{R@1} & mAP \\
     & & @1000 & & @1000 \\
    \midrule
    % \rowcolor{vlightgray}    
    % ProxyAnchor (PA) & 64.4 $\pm$ 0.3 & 33.2 $\pm$ 0.3 & 82.4 $\pm$ 0.4 & 34.2 $\pm$ 0.3\\
    % \rowcolor{vvlightgray}    
    PA + $\mathbb{NIR}$ & 66.0 $\pm$ 0.3 & 34.2 $\pm$ 0.2 & 85.2 $\pm$ 0.2 & 36.4 $\pm$ 0.2\\
    \midrule    
    Generate & 64.8 $\pm$ 0.4 & 33.0 $\pm$ 0.3 & 84.6 $\pm$ 0.6 & 35.9 $\pm$ 0.3 \\
    Reverse Match & 63.2 $\pm$ 0.1 & 31.0 $\pm$ 0.1 & 83.4 $\pm$ 0.1 & 32.8 $\pm$ 0.1 \\    
    Generate \& Match & 63.5 $\pm$ 0.2 & 31.6 $\pm$ 0.2 & 83.9 $\pm$ 0.4 & 35.7 $\pm$ 0.3 \\ 
    \bottomrule
    \end{tabular}}
    \label{tab:gen_ablations}
    \vspace{-5pt}
 \end{table}

\subsection{Self-regularization}
\label{subsec:exp_selfreg}
Finally, we study a natural extension to $\mathbb{NIR}$ by leveraging the generative process defined through our Normalizing Flow, which provides a translation from a probability density we can sample from (in this case just $\mathcal{N}(0, 1)$) to the representation space for a respective class $y$ (conditioned on $\rho_y$). Similar to \cite{dvml} or \cite{hardness-aware}, which saw benefits in generalisation through synthetic samples in sample-based DML, we investigate in Tab. \ref{tab:gen_ablations} whether sampling from the residual prior $\zeta \sim q(\cdot)$ and traversing the Normalizing Flow in reverse can generate synthetic samples that offer additional self-regularisation. More specifically, we investigate whether synthetic samples $\psi^s = \tau(\zeta|\rho)$ can be used in $\mathcal{L}_\text{PDML}({\psi^s}^X, \mathcal{P})$ to learn more generic proxies (``\textit{Generate}''), detach the synthetic samples $(^X)$. Similarly, we investigate whether the generated samples can be used to refine the quality of the Normalizing Flow by evaluating the matching quality with the learned proxies via $\mathcal{L}_\text{PDML}(\psi^s, \mathcal{P}^X)$ (``\textit{Match}''), as well as doing both steps jointly ($\mathcal{L}_\text{PDML}(\psi^s, \mathcal{P})$, ``\textit{Generate \& Match}).
%%%%
In its current setting, generalization and convergence suffer from artificial samples. Especially for the former, we see drops in performance from $66.0\%$ back down to $64.8\%$ when introducing artificial samples and even $63.2\%$ when applying reverse distribution matching. We hypothesize that this is due to the introduction of noisy samples especially in earlier training stages and the interdependence of $\mathbb{NIR}$ on the quality of the learned proxies. We believe a better adapted incorporation following e.g. a hardness-aware heuristic \cite{hardness-aware} could better leverage the benefits of such self-regularization. We leave this to future work to address.

%%%%%%%%%%%%%%%%%%%%%%%%%%%%%%%%%%%%%%%%%%%%%%%%%%%%%%%%%%%%%%%%%%%%%%%%
%%%%%%%%%%%%%%%%%%%%%%%%%%%%%%%%%%%%%%%%%%%%%%%%%%%%%%%%%%%%%%%%%%%%%%%%
\section{Conclusion}
\label{sec:conclusion}
This work proposes $\mathbb{NIR}$ - non-isotropy regularisation for proxy-based Deep Metric Learning (DML). $\mathbb{NIR}$ tackles the inherent problem of proxy-based objectives to resolve local structures and clusters to learn non-discriminative features that facilitate generalization, without influencing the superior convergence of proxy-based DML. $\mathbb{NIR}$ achieves this by refining the sample-distributional prior optimized in standard proxy-based DML through unique sample-proxy relation constraints. Extensive experiments support the idea of $\mathbb{NIR}$ and, besides the retention of fast convergence speeds, show significant improvements in the generalisation performance of proxy-based objectives, achieving competitive and state-of-the-art performance on all benchmarks.

\textbf{Limitations.} $\mathbb{NIR}$ relies on learning meaningful translations from (class-)proxies to respective samples. With high proxy count, the quality of these translations is impacted, evident in the performance on SOP. In addition, our current setting can not yet leverage the sample-generative process induced by the Normalizing Flows for additional regularization (which is also bottlenecked by high proxy-counts/low number of samples per class).

\textbf{Broader Impact.} Our work significantly benefits proxy-based DML. With fast convergence, this makes application in DML-driven domains s.a image \& video retrieval, but also face re-identification, very attractive. Especially for the latter more controversial application however, a potential for misuse given. However, while notable, improvements through $\mathbb{NIR}$ are not sufficient to drive a significant change in the societal usage in these domains.

\section*{Acknowledgements}
This work has been partially funded by the ERC (853489 - DEXIM) and by the DFG (2064/1 – Project number 390727645).
We thank the International Max Planck Research School for Intelligent Systems (IMPRS-IS) for supporting Karsten Roth. 
Karsten Roth further acknowledges
his membership in the European Laboratory for Learning
and Intelligent Systems (ELLIS) PhD program.

%%%%%%%%% REFERENCES
{\small
\bibliographystyle{ieee_fullname}
\bibliography{main}

\begin{thebibliography}{10}\itemsep=-1pt

\bibitem{cinn_imtoim}
Lynton Ardizzone, Jakob Kruse, Carsten L{\"{u}}th, Niels Bracher, Carsten
  Rother, and Ullrich K{\"{o}}the.
\newblock Conditional invertible neural networks for diverse image-to-image
  translation.
\newblock {\em CoRR}, abs/2105.02104, 2021.

\bibitem{cinn}
Lynton Ardizzone, Carsten L{\"u}th, Jakob Kruse, Carsten Rother, and Ullrich
  K{\"o}the.
\newblock Conditional invertible neural networks for guided image generation,
  2020.

\bibitem{rbf}
Piotr Bojanowski and Armand Joulin.
\newblock Unsupervised learning by predicting noise.
\newblock In {\em Proceedings of the 34th International Conference on Machine
  Learning - Volume 70}, ICML'17, page 517–526. JMLR.org, 2017.

\bibitem{Brattoli_2020_CVPR}
Biagio Brattoli, Joseph Tighe, Fedor Zhdanov, Pietro Perona, and Krzysztof
  Chalupka.
\newblock Rethinking zero-shot video classification: End-to-end training for
  realistic applications.
\newblock In {\em Proceedings of the IEEE/CVF Conference on Computer Vision and
  Pattern Recognition (CVPR)}, June 2020.

\bibitem{chen2020simple}
Ting Chen, Simon Kornblith, Mohammad Norouzi, and Geoffrey~Everest Hinton.
\newblock A simple framework for contrastive learning of visual
  representations.
\newblock 2020.

\bibitem{imagenet}
J. Deng, W. Dong, R. Socher, L.-J. Li, K. Li, and L. Fei-Fei.
\newblock {ImageNet: A Large-Scale Hierarchical Image Database}.
\newblock In {\em IEEE Conference on Computer Vision and Pattern Recognition
  (CVPR)}, 2009.

\bibitem{arcface}
J. {Deng}, J. {Guo}, N. {Xue}, and S. {Zafeiriou}.
\newblock Arcface: Additive angular margin loss for deep face recognition.
\newblock In {\em 2019 IEEE/CVF Conference on Computer Vision and Pattern
  Recognition (CVPR)}, pages 4685--4694, 2019.

\bibitem{nf_1}
Laurent Dinh, David Krueger, and Yoshua Bengio.
\newblock {NICE:} non-linear independent components estimation.
\newblock In Yoshua Bengio and Yann LeCun, editors, {\em 3rd International
  Conference on Learning Representations, {ICLR} 2015, San Diego, CA, USA, May
  7-9, 2015, Workshop Track Proceedings}, 2015.

\bibitem{realnvp}
Laurent Dinh, Jascha Sohl{-}Dickstein, and Samy Bengio.
\newblock Density estimation using real {NVP}.
\newblock In {\em 5th International Conference on Learning Representations,
  {ICLR} 2017, Toulon, France, April 24-26, 2017, Conference Track
  Proceedings}. OpenReview.net, 2017.

\bibitem{cinn_video}
Michael Dorkenwald, Timo Milbich, Andreas Blattmann, Robin Rombach,
  Konstantinos~G. Derpanis, and Bjorn Ommer.
\newblock Stochastic image-to-video synthesis using cinns.
\newblock In {\em Proceedings of the IEEE/CVF Conference on Computer Vision and
  Pattern Recognition (CVPR)}, pages 3742--3753, June 2021.

\bibitem{daml}
Yueqi Duan, Wenzhao Zheng, Xudong Lin, Jiwen Lu, and Jie Zhou.
\newblock Deep adversarial metric learning.
\newblock In {\em The IEEE Conference on Computer Vision and Pattern
  Recognition (CVPR)}, June 2018.

\bibitem{dullerud2022is}
Natalie Dullerud, Karsten Roth, Kimia Hamidieh, Nicolas Papernot, and Marzyeh
  Ghassemi.
\newblock Is fairness only metric deep? evaluating and addressing subgroup gaps
  in deep metric learning.
\newblock In {\em International Conference on Learning Representations}, 2022.

\bibitem{htl}
Weifeng Ge.
\newblock Deep metric learning with hierarchical triplet loss.
\newblock In {\em Proceedings of the European Conference on Computer Vision
  (ECCV)}, pages 269--285, 2018.

\bibitem{vmf_mix_2}
Siddharth Gopal and Yiming Yang.
\newblock Von mises-fisher clustering models.
\newblock In Eric~P. Xing and Tony Jebara, editors, {\em Proceedings of the
  31st International Conference on Machine Learning}, volume~32 of {\em
  Proceedings of Machine Learning Research}, pages 154--162, Bejing, China,
  22--24 Jun 2014. PMLR.

\bibitem{contrastive}
Raia Hadsell, Sumit Chopra, and Yann LeCun.
\newblock Dimensionality reduction by learning an invariant mapping.
\newblock In {\em Proceedings of the IEEE Conference on Computer Vision and
  Pattern Recognition}, 2006.

\bibitem{smartmining}
Ben Harwood, BG Kumar, Gustavo Carneiro, Ian Reid, Tom Drummond, et~al.
\newblock Smart mining for deep metric learning.
\newblock In {\em Proceedings of the IEEE International Conference on Computer
  Vision}, pages 2821--2829, 2017.

\bibitem{vmf_mix_1}
Md.~Abul Hasnat, Julien Bohn{\'{e}}, Jonathan Milgram, St{\'{e}}phane Gentric,
  and Liming Chen.
\newblock von mises-fisher mixture model-based deep learning: Application to
  face verification.
\newblock {\em CoRR}, abs/1706.04264, 2017.

\bibitem{moco}
Kaiming He, Haoqi Fan, Yuxin Wu, Saining Xie, and Ross Girshick.
\newblock Momentum contrast for unsupervised visual representation learning.
\newblock In {\em Proceedings of the IEEE/CVF Conference on Computer Vision and
  Pattern Recognition (CVPR)}, June 2020.

\bibitem{face_verfication_inthewild}
J. {Hu}, J. {Lu}, and Y. {Tan}.
\newblock Discriminative deep metric learning for face verification in the
  wild.
\newblock In {\em 2014 IEEE Conference on Computer Vision and Pattern
  Recognition}, 2014.

\bibitem{horde}
Pierre Jacob, David Picard, Aymeric Histace, and Edouard Klein.
\newblock Metric learning with horde: High-order regularizer for deep
  embeddings.
\newblock In {\em The IEEE Conference on Computer Vision and Pattern
  Recognition (CVPR)}, 2019.

\bibitem{rankmi}
Mete Kemertas, Leila Pishdad, Konstantinos~G. Derpanis, and Afsaneh Fazly.
\newblock Rankmi: A mutual information maximizing ranking loss.
\newblock In {\em Proceedings of the IEEE/CVF Conference on Computer Vision and
  Pattern Recognition (CVPR)}, June 2020.

\bibitem{khosla2020supervised}
Prannay Khosla, Piotr Teterwak, Chen Wang, Aaron Sarna, Yonglong Tian, Phillip
  Isola, Aaron Maschinot, Ce Liu, and Dilip Krishnan.
\newblock Supervised contrastive learning, 2020.

\bibitem{kim2020proxy}
Sungyeon Kim, Dongwon Kim, Minsu Cho, and Suha Kwak.
\newblock Proxy anchor loss for deep metric learning.
\newblock In {\em Proceedings of the IEEE/CVF Conference on Computer Vision and
  Pattern Recognition (CVPR)}, June 2020.

\bibitem{Kim_2021_CVPR}
Sungyeon Kim, Dongwon Kim, Minsu Cho, and Suha Kwak.
\newblock Embedding transfer with label relaxation for improved metric
  learning.
\newblock In {\em Proceedings of the IEEE/CVF Conference on Computer Vision and
  Pattern Recognition (CVPR)}, pages 3967--3976, June 2021.

\bibitem{abe}
Wonsik Kim, Bhavya Goyal, Kunal Chawla, Jungmin Lee, and Keunjoo Kwon.
\newblock Attention-based ensemble for deep metric learning.
\newblock In {\em Proceedings of the European Conference on Computer Vision
  (ECCV)}, 2018.

\bibitem{adam}
Diederik~P. Kingma and Jimmy Ba.
\newblock Adam: {A} method for stochastic optimization.
\newblock In Yoshua Bengio and Yann LeCun, editors, {\em 3rd International
  Conference on Learning Representations, {ICLR} 2015, San Diego, CA, USA, May
  7-9, 2015, Conference Track Proceedings}, 2015.

\bibitem{glow}
Durk~P Kingma and Prafulla Dhariwal.
\newblock Glow: Generative flow with invertible 1x1 convolutions.
\newblock In S. Bengio, H. Wallach, H. Larochelle, K. Grauman, N. Cesa-Bianchi,
  and R. Garnett, editors, {\em Advances in Neural Information Processing
  Systems}, volume~31. Curran Associates, Inc., 2018.

\bibitem{memvir}
Byungsoo Ko, Geonmo Gu, and Han-Gyu Kim.
\newblock Learning with memory-based virtual classes for deep metric learning.
\newblock In {\em Proceedings of the IEEE/CVF International Conference on
  Computer Vision (ICCV)}, pages 11792--11801, October 2021.

\bibitem{kobyzev2019normalizing}
Ivan Kobyzev, Simon Prince, and Marcus~A. Brubaker.
\newblock Normalizing flows: An introduction and review of current methods.
\newblock 2019.
\newblock cite arxiv:1908.09257Comment: Updated version, currently under review
  in a journal.

\bibitem{cars196}
Jonathan Krause, Michael Stark, Jia Deng, and Li Fei-Fei.
\newblock 3d object representations for fine-grained categorization.
\newblock In {\em Proceedings of the IEEE International Conference on Computer
  Vision Workshops}, pages 554--561, 2013.

\bibitem{dvml}
Xudong Lin, Yueqi Duan, Qiyuan Dong, Jiwen Lu, and Jie Zhou.
\newblock Deep variational metric learning.
\newblock In {\em The European Conference on Computer Vision (ECCV)}, September
  2018.

\bibitem{sphereface}
Weiyang Liu, Yandong Wen, Zhiding Yu, Ming Li, Bhiksha Raj, and Le Song.
\newblock Sphereface: Deep hypersphere embedding for face recognition.
\newblock {\em IEEE Conference on Computer Vision and Pattern Recognition
  (CVPR)}, 2017.

\bibitem{torchvision}
S\'{e}bastien Marcel and Yann Rodriguez.
\newblock Torchvision the machine-vision package of torch.
\newblock In {\em Proceedings of the 18th ACM International Conference on
  Multimedia}, MM '10, page 1485–1488, New York, NY, USA, 2010. Association
  for Computing Machinery.

\bibitem{milbich2020diva}
Timo Milbich, Karsten Roth, Homanga Bharadhwaj, Samarth Sinha, Yoshua Bengio,
  Bj{\"{o}}rn Ommer, and Joseph~Paul Cohen.
\newblock Diva: Diverse visual feature aggregation for deep metric learning.
\newblock {\em CoRR}, abs/2004.13458, 2020.

\bibitem{milbich2020sharing}
T. {Milbich}, K. {Roth}, B. {Brattoli}, and B. {Ommer}.
\newblock Sharing matters for generalization in deep metric learning.
\newblock {\em IEEE Transactions on Pattern Analysis and Machine Intelligence},
  pages 1--1, 2020.

\bibitem{milbich2021characterizing}
Timo Milbich, Karsten Roth, Samarth Sinha, Ludwig Schmidt, Marzyeh Ghassemi,
  and Bj{\"o}rn Ommer.
\newblock Characterizing generalization under out-of-distribution shifts in
  deep metric learning.
\newblock In A. Beygelzimer, Y. Dauphin, P. Liang, and J.~Wortman Vaughan,
  editors, {\em Advances in Neural Information Processing Systems}, 2021.

\bibitem{pretextmisra}
Ishan Misra and Laurens van~der Maaten.
\newblock Self-supervised learning of pretext-invariant representations.
\newblock In {\em 2020 {IEEE/CVF} Conference on Computer Vision and Pattern
  Recognition, {CVPR} 2020, Seattle, WA, USA, June 13-19, 2020}, pages
  6706--6716. {IEEE}, 2020.

\bibitem{proxynca}
Yair Movshovitz-Attias, Alexander Toshev, Thomas~K Leung, Sergey Ioffe, and
  Saurabh Singh.
\newblock No fuss distance metric learning using proxies.
\newblock In {\em Proceedings of the IEEE International Conference on Computer
  Vision}, pages 360--368, 2017.

\bibitem{musgrave2020metric}
Kevin Musgrave, Serge~J. Belongie, and Ser{-}Nam Lim.
\newblock A metric learning reality check.
\newblock {\em CoRR}, abs/2003.08505, 2020.

\bibitem{lifted}
Hyun Oh~Song, Yu Xiang, Stefanie Jegelka, and Silvio Savarese.
\newblock Deep metric learning via lifted structured feature embedding.
\newblock In {\em Proceedings of the IEEE Conference on Computer Vision and
  Pattern Recognition}, pages 4004--4012, 2016.

\bibitem{nf_2}
George Papamakarios, Theo Pavlakou, and Iain Murray.
\newblock Masked autoregressive flow for density estimation.
\newblock In I. Guyon, U.~V. Luxburg, S. Bengio, H. Wallach, R. Fergus, S.
  Vishwanathan, and R. Garnett, editors, {\em Advances in Neural Information
  Processing Systems}, volume~30. Curran Associates, Inc., 2017.

\bibitem{pytorch}
Adam Paszke, Sam Gross, Soumith Chintala, Gregory Chanan, Edward Yang, Zachary
  DeVito, Zeming Lin, Alban Desmaison, Luca Antiga, and Adam Lerer.
\newblock Automatic differentiation in pytorch.
\newblock In {\em NIPS-W}, 2017.

\bibitem{softriple}
Qi Qian, Lei Shang, Baigui Sun, Juhua Hu, Hao Li, and Rong Jin.
\newblock Softtriple loss: Deep metric learning without triplet sampling.
\newblock In {\em Proceedings of the IEEE/CVF International Conference on
  Computer Vision (ICCV)}, October 2019.

\bibitem{normflows}
Danilo~Jimenez Rezende and Shakir Mohamed.
\newblock Variational inference with normalizing flows.
\newblock In {\em Proceedings of the 32nd International Conference on
  International Conference on Machine Learning - Volume 37}, ICML'15, page
  1530–1538. JMLR.org, 2015.

\bibitem{cinn_translate}
Robin Rombach, Patrick Esser, and Bjorn Ommer.
\newblock Network-to-network translation with conditional invertible neural
  networks.
\newblock In H. Larochelle, M. Ranzato, R. Hadsell, M.~F. Balcan, and H. Lin,
  editors, {\em Advances in Neural Information Processing Systems}, volume~33,
  pages 2784--2797. Curran Associates, Inc., 2020.

\bibitem{mic}
Karsten Roth, Biagio Brattoli, and Bjorn Ommer.
\newblock Mic: Mining interclass characteristics for improved metric learning.
\newblock In {\em Proceedings of the IEEE International Conference on Computer
  Vision}, pages 8000--8009, 2019.

\bibitem{roth2020pads}
Karsten Roth, Timo Milbich, and Bjorn Ommer.
\newblock Pads: Policy-adapted sampling for visual similarity learning.
\newblock In {\em Proceedings of the IEEE/CVF Conference on Computer Vision and
  Pattern Recognition (CVPR)}, June 2020.

\bibitem{s2sd}
Karsten Roth, Timo Milbich, Bjorn Ommer, Joseph~Paul Cohen, and Marzyeh
  Ghassemi.
\newblock Simultaneous similarity-based self-distillation for deep metric
  learning.
\newblock In Marina Meila and Tong Zhang, editors, {\em Proceedings of the 38th
  International Conference on Machine Learning}, volume 139 of {\em Proceedings
  of Machine Learning Research}, pages 9095--9106. PMLR, 18--24 Jul 2021.

\bibitem{roth2020revisiting}
Karsten Roth, Timo Milbich, Samarth Sinha, Prateek Gupta, Bjorn Ommer, and
  Joseph~Paul Cohen.
\newblock Revisiting training strategies and generalization performance in deep
  metric learning.
\newblock In Hal~Daumé III and Aarti Singh, editors, {\em Proceedings of the
  37th International Conference on Machine Learning}, volume 119 of {\em
  Proceedings of Machine Learning Research}, pages 8242--8252. PMLR, 13--18 Jul
  2020.

\bibitem{samesame}
Marco Rudolph, Bastian Wandt, and Bodo Rosenhahn.
\newblock Same same but differnet: Semi-supervised defect detection with
  normalizing flows.
\newblock In {\em Proceedings of the IEEE/CVF Winter Conference on Applications
  of Computer Vision (WACV)}, pages 1907--1916, January 2021.

\bibitem{Sanakoyeu_2019_CVPR}
Artsiom Sanakoyeu, Vadim Tschernezki, Uta Buchler, and Bjorn Ommer.
\newblock Divide and conquer the embedding space for metric learning.
\newblock In {\em The IEEE Conference on Computer Vision and Pattern
  Recognition (CVPR)}, 2019.

\bibitem{semihard}
Florian Schroff, Dmitry Kalenichenko, and James Philbin.
\newblock Facenet: A unified embedding for face recognition and clustering.
\newblock In {\em Proceedings of the IEEE conference on computer vision and
  pattern recognition}, pages 815--823, 2015.

\bibitem{seidenschwarz2021graphdml}
Jenny~Denise Seidenschwarz, Ismail Elezi, and Laura Leal-Taix{\'e}.
\newblock Learning intra-batch connections for deep metric learning.
\newblock In Marina Meila and Tong Zhang, editors, {\em Proceedings of the 38th
  International Conference on Machine Learning}, volume 139 of {\em Proceedings
  of Machine Learning Research}, pages 9410--9421. PMLR, 18--24 Jul 2021.

\bibitem{npairs}
Kihyuk Sohn.
\newblock Improved deep metric learning with multi-class n-pair loss objective.
\newblock In {\em Advances in Neural Information Processing Systems}, pages
  1857--1865, 2016.

\bibitem{circle}
Yifan Sun, Changmao Cheng, Yuhan Zhang, Chi Zhang, Liang Zheng, Zhongdao Wang,
  and Yichen Wei.
\newblock Circle loss: A unified perspective of pair similarity optimization.
\newblock In {\em Proceedings of the IEEE/CVF Conference on Computer Vision and
  Pattern Recognition (CVPR)}, June 2020.

\bibitem{proxyncapp}
Eu~Wern Teh, Terrance DeVries, and Graham~W. Taylor.
\newblock Proxynca++: Revisiting and revitalizing proxy neighborhood component
  analysis.
\newblock In Andrea Vedaldi, Horst Bischof, Thomas Brox, and Jan{-}Michael
  Frahm, editors, {\em Computer Vision - {ECCV} 2020 - 16th European
  Conference, Glasgow, UK, August 23-28, 2020, Proceedings, Part {XXIV}},
  volume 12369 of {\em Lecture Notes in Computer Science}, pages 448--464.
  Springer, 2020.

\bibitem{cub200-2011}
C. Wah, S. Branson, P. Welinder, P. Perona, and S. Belongie.
\newblock The caltech-ucsd birds-200-2011 dataset.
\newblock Technical Report CNS-TR-2011-001, California Institute of Technology,
  2011.

\bibitem{eq_1}
Feng Wang, Xiang Xiang, Jian Cheng, and Alan~Loddon Yuille.
\newblock Normface: L<sub>2</sub> hypersphere embedding for face verification.
\newblock In {\em Proceedings of the 25th ACM International Conference on
  Multimedia}, MM '17, page 1041–1049, New York, NY, USA, 2017. Association
  for Computing Machinery.

\bibitem{cosface}
Hao Wang, Yitong Wang, Zheng Zhou, Xing Ji, Dihong Gong, Jingchao Zhou, Zhifeng
  Li, and Wei Liu.
\newblock Cosface: Large margin cosine loss for deep face recognition, 2018.

\bibitem{wang2020understanding}
Tongzhou Wang and Phillip Isola.
\newblock Understanding contrastive representation learning through alignment
  and uniformity on the hypersphere.
\newblock {\em arXiv preprint arXiv:2005.10242}, 2020.

\bibitem{multisimilarity}
Xun Wang, Xintong Han, Weilin Huang, Dengke Dong, and Matthew~R. Scott.
\newblock Multi-similarity loss with general pair weighting for deep metric
  learning.
\newblock In {\em Proceedings of the IEEE/CVF Conference on Computer Vision and
  Pattern Recognition (CVPR)}, June 2019.

\bibitem{timm}
Ross Wightman.
\newblock Pytorch image models.
\newblock \url{https://github.com/rwightman/pytorch-image-models}, 2019.

\bibitem{margin}
Chao-Yuan Wu, R Manmatha, Alexander~J Smola, and Philipp Krahenbuhl.
\newblock Sampling matters in deep embedding learning.
\newblock In {\em Proceedings of the IEEE International Conference on Computer
  Vision}, pages 2840--2848, 2017.

\bibitem{epshn}
Hong Xuan, Abby Stylianou, and Robert Pless.
\newblock Improved embeddings with easy positive triplet mining.
\newblock In {\em Proceedings of the IEEE/CVF Winter Conference on Applications
  of Computer Vision (WACV)}, March 2020.

\bibitem{eq_2}
Dingyi Zhang, Yingming Li, and Zhongfei Zhang.
\newblock Deep metric learning with spherical embedding.
\newblock In {\em NeurIPS}, 2020.

\bibitem{hardness-aware}
Wenzhao Zheng, Zhaodong Chen, Jiwen Lu, and Jie Zhou.
\newblock Hardness-aware deep metric learning.
\newblock {\em The IEEE Conference on Computer Vision and Pattern Recognition
  (CVPR)}, 2019.

\bibitem{Zheng_2021_CVPR_compositional}
Wenzhao Zheng, Chengkun Wang, Jiwen Lu, and Jie Zhou.
\newblock Deep compositional metric learning.
\newblock In {\em Proceedings of the IEEE/CVF Conference on Computer Vision and
  Pattern Recognition (CVPR)}, pages 9320--9329, June 2021.

\bibitem{drml}
Wenzhao Zheng, Borui Zhang, Jiwen Lu, and Jie Zhou.
\newblock Deep relational metric learning.
\newblock In {\em Proceedings of the IEEE/CVF International Conference on
  Computer Vision (ICCV)}, pages 12065--12074, October 2021.

\bibitem{Zhu2020graphdml}
Yuehua Zhu, Muli Yang, Cheng Deng, and Wei Liu.
\newblock Fewer is more: A deep graph metric learning perspective using fewer
  proxies.
\newblock In H. Larochelle, M. Ranzato, R. Hadsell, M.~F. Balcan, and H. Lin,
  editors, {\em Advances in Neural Information Processing Systems}, volume~33,
  pages 17792--17803. Curran Associates, Inc., 2020.

\bibitem{dml_inversion}
Roland~S. Zimmermann, Yash Sharma, Steffen Schneider, Matthias Bethge, and
  Wieland Brendel.
\newblock Contrastive learning inverts the data generating process.
\newblock In Marina Meila and Tong Zhang, editors, {\em Proceedings of the 38th
  International Conference on Machine Learning}, volume 139 of {\em Proceedings
  of Machine Learning Research}, pages 12979--12990. PMLR, 18--24 Jul 2021.

\end{thebibliography}
}

\end{document}